\documentclass{article}
\usepackage{iclr2027_conference,times}
\iclrfinalcopy
\usepackage[T1]{fontenc}
\usepackage[utf8]{inputenc}
\usepackage{amsmath,amssymb,mathtools}
\usepackage{graphicx,booktabs,array,tabularx,multirow}
\usepackage{caption}
\usepackage{xcolor,microtype}
\usepackage{algorithm,algpseudocode}
\usepackage{enumitem,etoolbox}
\usepackage[most]{tcolorbox}
\usepackage{tikz}
\usetikzlibrary{arrows.meta,positioning,calc,fit,backgrounds}
\usepackage{hyperref,url}
\hypersetup{colorlinks=true,linkcolor=black,citecolor=black,urlcolor=black,
 pdftitle={Recursive Code World Models},pdfauthor={Zhiqi Li, Yuxuan Liao, Bo Zhu}}
\definecolor{draftgray}{HTML}{536071}
\newif\ifsubmission
\submissionfalse % Set true only after evidence, anonymity, and page-limit checks.
\ifsubmission\else
\makeatletter
\patchcmd{\@maketitle}{Paper under double-blind review}{Preprint}{}{}
\makeatother
\fi
\newcommand{\method}{\textsc{RCWM}}
\newcommand{\rsp}{\textsc{RSP}}

\newcommand{\Render}{\mathcal{R}}
\newcommand{\Exec}{\operatorname{Exec}}
\newcommand{\Compose}{\operatorname{Compose}}
\newcommand{\Observe}{\operatorname{Observe}}

\newcommand{\code}[1]{\texttt{#1}}
\newcolumntype{Y}{>{\raggedright\arraybackslash}X}
\setlist[itemize]{leftmargin=*,itemsep=1pt,topsep=3pt}
\setlist[enumerate]{leftmargin=*,itemsep=1pt,topsep=3pt}
\title{Recursive Code World Models:\\Building Complex Worlds through\\Recursive Scene Programs}
\author{
Zhiqi Li\thanks{Correspondence: \texttt{zli3167@gatech.edu}}
\quad Yuxuan Liao
\quad Bo Zhu\\
Georgia Institute of Technology
}
\begin{document}
\maketitle
\vspace{-2.1em}
\noindent
{\small
\textbf{Project:}~
\href{https://zhiqili-cg.github.io/RecursiveCWM}
{\texttt{https://zhiqili-cg.github.io/RecursiveCWM}}\\
\hspace{1.5em}
\textbf{Code:}~
\href{https://github.com/ZhiqiLi-CG/RecursiveCWM_code}
{\texttt{https://github.com/ZhiqiLi-CG/RecursiveCWM\_code}}
}
% Teaser: full-width banner directly under the author block, before the abstract (SIGGRAPH-style placement).
% The image is allowed to run slightly into the margins; the caption stays at text width.

\begin{center}
\makebox[\linewidth][c]{\includegraphics[width=1.12\linewidth]{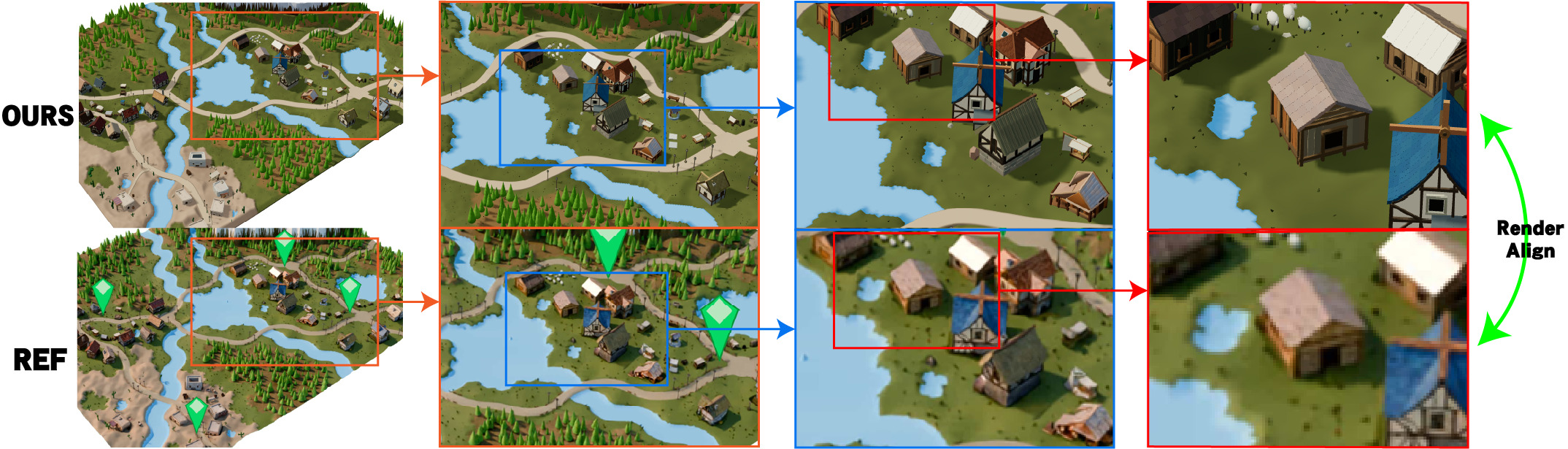}}
\vspace{-0.2em}
\begin{minipage}{\linewidth}
\captionsetup{type=figure,font=small,labelfont=bf,skip=2pt,justification=justified,singlelinecheck=false}
\captionof{figure}{For image-to-3D code world reconstruction, Recursive Code World Models
use recursive global--local--global construction to progressively refine
executable 3D scene programs, producing increasingly accurate local geometry
and appearance while preserving whole-scene consistency.}
\label{fig:teaser}
\end{minipage}
\end{center}
\vspace{0.3em}
\ifsubmission\else
\lhead{Preprint \hfill September 2026}
\raggedbottom % Avoid stretched float gaps in the working draft; submission retains template default.
\fi
\begin{abstract}
Code world models represent worlds as executable programs, but this representation alone does not determine how to construct a complex world. We introduce \emph{Recursive Code World Models} (\method), a framework for reconstructing complex 3D worlds in code from a single reference image. \method{} couples a \emph{Recursive Scene Program} (\rsp) representation with a construction solver that recursively calls itself. An \rsp{} represents the executable world as compositional scene code, while each solver call follows the same complete process: establish the whole, recursively reconstruct unresolved parts, and revisit the whole to refine their composition. This global--local--global recursion gives fine-scale structures their own perception-and-editing loops while preserving scene-wide geometry and relationships. Reference-aligned views propagate a shared camera projection across levels, while parent revisitation addresses boundaries, spatial relations, and shared errors that emerge after local refinement. A vision-language coding agent directly compares reference images with scene renders to guide refinement, recursive descent, and return. Across complex scenes, \method{} outperforms prior code-based image-to-scene reconstruction methods. Ablation studies further support the benefits of recursive construction and suggest that deeper calls can improve finer-scale reconstruction. \method{} provides a recursive construction principle for building complex executable worlds from visual evidence.
\end{abstract}
\section{Introduction}
\label{sec:intro}
Code offers an executable and compositional representation of a world, exposing its structure and state for direct inspection and manipulation rather than leaving them implicit in generated pixels \citep{chen2026codeworld}. By expressing geometry, materials, and spatial relationships as program elements, scene code supports targeted editing, component reuse, and integration with physical simulation \citep{yin2026viga}. Recent vision-language coding agents have begun to realize this potential by reconstructing scenes from images through program generation and render-based visual inspection \citep{yin2026viga,he2026thinking,img2threejs}. Yet constructing a complex world requires more than an expressive representation: it requires resolving structures at multiple scales while maintaining the relationships among its parts. This raises a fundamental question: \emph{how should a coding model organize its computation when the world contains more structure than a single reconstruction trajectory can resolve?}

Consider rebuilding a town from an image. Streets and terrain define its layout; buildings contain facades, and facades contain windows, signs, and ornaments. A scene-wide reconstruction establishes a shared spatial context, but repeated whole-scene reviews may still overlook these small structures. Giving them focused reconstruction tasks makes their details easier to inspect and refine, yet local edits are not independent: a sign may overlap a neighboring window incorrectly, or a refined bridge may no longer meet the riverbanks. These conflicts can emerge even from a coherent initial layout and may only become apparent when the refined parts are inspected together. The same tension arises at every scale, from buildings within a town to windows and signs within a facade.

Our central claim is that \textbf{complex-world construction should be recursive}. A part is not merely an asset to generate and insert. It is a subworld that deserves its own complete reconstruction process, with inherited context, focused visual evidence, and a return to the whole that contains it. The essential recurrence is
\begin{equation}
\underbrace{
  \boxed{
    \begin{gathered}
      \text{\bfseries Level } k\\[-1pt]
      \text{Establish the whole}
    \end{gathered}
  }
  \;\xrightarrow{\text{recurse}}\;
  \boxed{
    \begin{gathered}
      \text{\bfseries Level } k+1\\[-1pt]
      \text{Call }\mathcal{F}\text{ on each part}
    \end{gathered}
  }
  \;\xrightarrow{\text{return}}\;
  \boxed{
    \begin{gathered}
      \text{\bfseries Level } k\\[-1pt]
      \text{Revisit the whole}
    \end{gathered}
  }
}_{\mathcal{F}\,:\;\text{the same solver at every level}}
\label{eq:recursion-intuition}
\end{equation}
The first whole establishes a shared spatial context for local construction. The second inspects and refines the assembled whole, correcting inconsistencies among its parts and their relationships. Crucially, the same complete cycle applies within a building, a facade, or a terrain region. This is not a fixed scene--object pipeline followed by a final cleanup, but a construction process that recursively calls itself: each subworld establishes its own whole, reconstructs its unresolved parts through the same solver, and refines their composition before returning to its parent.

\begin{figure}[t]
\centering
\vspace{-14mm}
\includegraphics[width=\linewidth]{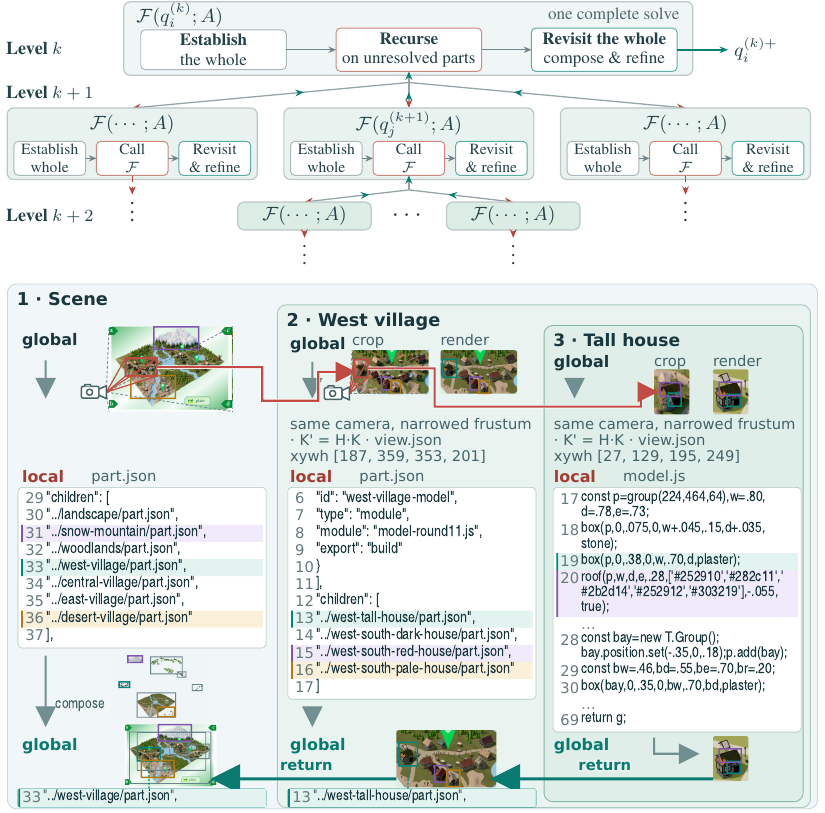}
\vspace{-4mm}
\caption{\textbf{Recursive Code World Models.}
\emph{Top:} the recursive construction process. At level $k$, the
solver $\mathcal F$ establishes the current whole, calls the same
solver on unresolved subworlds at level $k+1$, and revisits and
refines the whole after their results return. Each child can in turn
make further calls at level $k+2$ and beyond, producing a recursive
call tree.
\emph{Bottom:} one recorded construction path on
\emph{medieval-village}: the scene calls a solver for the west village,
which in turn calls a solver for the tall house. Each level receives
a matched reference crop and camera view, returns its constructed
subprogram to the parent, and is then revisited in the parent
composition.}
\vspace{-5mm}
\label{fig:overview}
\end{figure}

We introduce \emph{Recursive Code World Models} (\method) and their output representation, \emph{Recursive Scene Programs} (\rsp{}s). An \rsp{} represents an executable world as compositional scene code organized over nested subworlds. \method{} constructs this representation through a complete recursive process. At each node, a vision-language coding agent establishes a whole, invokes the same solver on unresolved visual subproblems, and then inspects and refines their composition. Each child receives focused reference evidence and its own perception-and-editing loop, while inheriting the parent's camera projection and spatial conventions. Parent revisitation addresses boundaries, spatial relationships, and shared causes of error that isolated local reviews may overlook. Together, these mechanisms allow local detail and cross-part consistency to be addressed at every level, with visual feedback guiding further descent where finer structure remains unresolved. Recursion determines which subproblems receive a complete solve, whereas parallelism only changes how compatible branches are scheduled. Our contributions are threefold:
\begin{itemize}
    \item \textbf{Recursive Scene Programs.}
    We introduce an executable scene representation that organizes complex worlds as compositional subworld programs with explicit child references, editable structure, and shared dependencies.
    \item \textbf{Recursive Code World Construction.}
    We introduce a complete recursive construction process in which the same visual solver establishes each subworld, recursively reconstructs its unresolved parts, and revisits their composition after return, combining inherited observation geometry, visually guided recursive calls, and active parent-level refinement.
    \item \textbf{Reference-driven reconstruction and evaluation.}
    We instantiate the framework with a parameterized Three.js compiler and evaluate whole-scene and local reconstruction against image-to-scene-program baselines using the same base model. Ablation studies examine the roles of construction order, recursive depth, and parent revisitation under the same local visual access.
\end{itemize}
% Method replacement: output representation followed by an algorithm-first
% construction subsection. Uses existing macros \Exec, \Render, \Observe,
% and \Compose. Required packages: amsmath, graphicx, algorithm, algpseudocode.
% The [H] algorithm placement is provided by float (loaded by algorithm).
% Uses the existing project figure figures/visual_interfaces.pdf.

\section{Recursive Code World Models}
\label{sec:method}

\subsection{Executable Recursive Scene Programs}
\label{sec:representation}

Given a single reference image $I^{\star}$, our goal is to reconstruct
its depicted world as an executable, parameterized scene program $P$.
We represent this output as a \emph{Recursive Scene Program} (RSP),
containing source procedures, editable parameters, component references,
and the shared dependencies needed to construct the world. Let $\Exec$
denote program execution, $S$ the resulting 3D scene, $\Render$ the
rendering operation, and $\widehat I$ its output image. With a
reference-view camera configuration $\kappa$, execution and rendering
give
\begin{equation}
S=\Exec(P), \qquad \widehat I=\Render(S;\kappa).
\label{eq:task}
\end{equation}
The program specifies geometry, materials, lighting, and component
placement; $\kappa$ specifies the camera pose, projection, and viewport
dimensions. We seek a program whose rendered layout, local appearance,
and spatial relationships agree with the visible evidence in
$I^{\star}$. The delivered code supports direct execution, parameter
editing, and rendering from additional views.

\paragraph{Compositional program structure.}
The world program is organized into nested subworlds, each representing
an object, an interacting assembly, or a continuous surface region.
A node $u_i^{(k)}$ denotes a subworld at depth $k$ with program
$P_i^{(k)}$, where $i$ is unique across the hierarchy and the root is
$u_0^{(0)}$. The set $\operatorname{Children}(i)$ contains its direct
child identifiers, so $j\in\operatorname{Children}(i)$ identifies a
child $u_j^{(k+1)}$ with program $P_j^{(k+1)}$.
Each parent program contains local construction procedures, editable
parameters, and references to these child programs, together with
the rules for their placement and interfaces.
We write $P_i^{(k)}[j]$ for fetching a child program: it follows the
reference labeled $j$ in the parent code and retrieves the
corresponding child code, giving $P_i^{(k)}[j]=P_j^{(k+1)}$.
For selected children, let $Q_j$ denote a candidate replacement for
child program $P_j^{(k+1)}$. The operation
$\Compose(P_i^{(k)},\{j\mapsto Q_j\})$ produces an updated parent
program whose selected child references point to $Q_j$, preserving
the parent's local code, placement rules, and remaining children.
Shared generators and assets are accessed through code references,
and spatial relationships can connect components across branches.
The root program $P = P_0^{(0)}$ forms the executable world, which can be executed in Three.js to construct the complete world.

\subsection{Recursive World Construction}
\label{sec:construction}

\begin{algorithm}[H]
\caption{Recursive construction of a scene program}
\label{alg:rcwm}
\begin{algorithmic}[1]
\Require Reference image $I^{\star}$; total budget $b$; coding agent $A$
\Ensure Executable program $P$ and reference-view camera $\kappa$
\State Initialize the root state $q_0^{(0)}$ and identity context $\mathcal C_0^{(0)}$.
  \Comment{\S\ref{sec:construction-state}}
  \label{algline:initialize}
\State $q_0^{(0)+}\gets\mathcal F(q_0^{(0)},\mathcal C_0^{(0)};A)$
  \label{algline:root-call}
\State \Return the program $P$ and camera $\kappa$ from $q_0^{(0)+}$.
  \Comment{\S\ref{sec:revisit}}
  \label{algline:deliver}
\Statex
\Function{$\mathcal F$}{$q_i^{(k)},\mathcal C_i^{(k)};A$}
  \While{budget remains in $q_i^{(k)}$}
    \Statex \textit{1. Establish the whole}
      \hfill \Comment{\S\ref{sec:establish}}
    \State Compare the current contextual render with its reference using $A$.
      \label{algline:observe}
    \State Refine the layout, shared structures, and component interfaces.
      \label{algline:establish}
    \Statex \textit{2. Prepare and recursively construct the parts}
      \hfill \Comment{\S\ref{sec:descent}}
    \State Select unresolved children $\mathcal U_i^{(k)}$;
      create or reopen their programs.
      \label{algline:select}
    \State Prepare child states $q_j^{(k+1)}$ and contexts
      $\mathcal C_j^{(k+1)}$ from the updated parent snapshot, with code,
      matched views, and allocated budgets.
      \label{algline:prepare}
    \If{the budget in $q_i^{(k)}$ is exhausted}
      \State \Return $q_i^{(k)}$
    \EndIf
    \ForAll{$j\in\mathcal U_i^{(k)}$}
      \State $q_j^{(k+1)+}\gets\mathcal F(q_j^{(k+1)},\mathcal C_j^{(k+1)};A)$
        \Comment{The same complete solver}
        \label{algline:recurse}
    \EndFor
    \Statex \textit{3. Compose, refine, and return}
      \hfill \Comment{\S\ref{sec:revisit}}
    \State Integrate returned child programs and update the parent's budget.
      \label{algline:compose}
    \If{the budget in $q_i^{(k)}$ is exhausted}
      \State \Return $q_i^{(k)}$
    \EndIf
    \State Render the assembled whole and refine cross-part relationships.
      \label{algline:refine}
    \State Review objects, relationships, and contact boundaries together.
      \label{algline:review}
    \State Update the refined program and remaining budget in $q_i^{(k)}$.
    \If{a complete review finds no new actionable discrepancy}
      \State \Return $q_i^{(k)}$
        \label{algline:visual-return}
    \EndIf
  \EndWhile
  \State \Return $q_i^{(k)}$
    \label{algline:budget-return}
\EndFunction
\end{algorithmic}
\end{algorithm}

We construct the scene program with a recursive solver $\mathcal F$
driven by a frozen vision-language coding agent $A$. Each call
follows three stages: \emph{establish the whole}, \emph{prepare and
recursively construct the parts}, and \emph{compose, refine, and
return}. At depth $k$, the solver establishes the subworld's layout,
gives selected components complete solves at depth $k+1$, and returns
to depth $k$ to refine their composition. Every child follows this
same complete process. Throughout construction, $P_i^{(k)}$ denotes
the current version of its component code, and visual feedback guides
local edits, further child calls, and the return to the parent.

In Algorithm~\ref{alg:rcwm}, $b$ is the total inference budget and
$q_i^{(k)}$ contains the current component code, reference view,
matched camera, and remaining budget. The inherited assembly function
$\mathcal C_i^{(k)}$ places candidate component code in the surrounding
world using the parent program snapshot. The selected unresolved child identifiers
form $\mathcal U_i^{(k)}\subseteq\operatorname{Children}(i)$, and
superscript $+$ marks a returned version. The state and context are
defined in Sections~\ref{sec:construction-state} and~\ref{sec:establish}.

Algorithm~\ref{alg:rcwm} gives the complete workflow from initialization
to the final program; the single instruction that realizes it at every node of our
implementation is reproduced in Appendix~\ref{app:solver}. Each call receives its local working state and
inherited context as separate inputs:
\begin{equation}
q_i^{(k)+}=\mathcal F(q_i^{(k)},\mathcal C_i^{(k)};A).
\label{eq:solver-interface}
\end{equation}
The following subsections define the state and expand the stages indicated in the
algorithm.

\subsubsection{Construction state and initialization}
\label{sec:construction-state}

Line~\ref{algline:initialize} of Algorithm~\ref{alg:rcwm} initializes
the root. Each call maintains a local working state during construction,
combining its candidate code with the reference evidence, camera, and
resources used to revise it, which are constructed in Section~\ref{sec:descent}. 

\paragraph{Working state.}
Let $I_i^{(k)}$ be the call's reference view,
$\kappa_i^{(k)}$ its matched camera, and $b_i^{(k)}$ its remaining
inference budget, including descendant calls. The state is
\begin{equation}
q_i^{(k)}=
\left(P_i^{(k)},I_i^{(k)},\kappa_i^{(k)},b_i^{(k)}\right).
\label{eq:state}
\end{equation}
The program field holds the evolving component code; the other fields
specify the evidence and resources for its construction. Repeated calls
on a component retain its identifier and depth, with their individual
visits recorded in the construction trace.

\paragraph{Root initialization.}
The procedure $\operatorname{Initialize}$ creates a seed program,
establishes world-coordinate and unit conventions, and calibrates the
reference camera through full-frame visual inspection. It produces
\begin{equation}
\begin{aligned}
\left(P_0^{(0)},\kappa_0^{(0)},b_0^{(0)}\right)
&=\operatorname{Initialize}(I^{\star},b;A),\\
q_0^{(0)}
&=\left(P_0^{(0)},I_0^{(0)}=I^{\star},\kappa_0^{(0)},b_0^{(0)}\right).
\end{aligned}
\label{eq:initialization}
\end{equation}
Here $b_0^{(0)}$ is the budget remaining
after initialization.  Child calls inherit the root's coordinate
conventions and derive their cameras from its projection. The root
controls camera revisions; each revision is followed by refreshed
observations and affected child states before further descent. (see Section~\ref{sec:descent}).

\subsubsection{Establish the whole}
\label{sec:establish}
\label{sec:observation}

Lines~\ref{algline:observe}--\ref{algline:establish} of
Algorithm~\ref{alg:rcwm} establish the current subworld's layout and
the interfaces among its parts through visual comparison. The agent
renders the subworld within its inherited scene context, which
provides the surrounding geometry, shared structures, and placement
rules. Comparing this render with the reference guides revisions
to the subworld and establishes a shared spatial arrangement for
subsequent child work.

\paragraph{Observe and refine.}
Let $\mathcal C_i^{(k)}[\,\cdot\,]$ denote the inherited program
context. For candidate component code $Q$, the expression
$\mathcal C_i^{(k)}[Q]$ assembles a complete world program by placing
$Q$ in the current component's slot while preserving the surrounding
scene code. The observation operator $\Observe$ executes this
assembled program and pairs its render with the reference:
\begin{equation}
\Observe(q_i^{(k)},\mathcal C_i^{(k)})=
\left[
I_i^{(k)},\;
\Render\!\left(
  \Exec\!\left(\mathcal C_i^{(k)}[P_i^{(k)}]\right);
  \kappa_i^{(k)}
\right)
\right].
\label{eq:observation}
\end{equation}
The agent directly examines this pair, edits $P_i^{(k)}$, and
renders the updated program for another comparison. Each pass can
contain several observation--edit--render iterations. A building
establishes its facade arrangement before individual details are
reconstructed; a terrain region establishes the surfaces and
interfaces needed by local structures. Shared geometry is authored
at a scope that contains its dependencies.

\paragraph{Context defined by the parent program.}
The inherited context is constructed from the parent program when
the child call is prepared. Let $\bar P_i^{(k)}$ denote the parent
program snapshot after its whole-construction and child-preparation
phase. For each child $j\in\operatorname{Children}(i)$, the assembly
function is defined recursively as
\begin{equation}
\begin{aligned}
\mathcal C_0^{(0)}[Q]
&\mathrel{:=} Q,\\
\mathcal C_j^{(k+1)}[Q]
&\mathrel{:=} \mathcal C_i^{(k)}\!\left[
\Compose\!\left(\bar P_i^{(k)},\{j\mapsto Q\}\right)
\right].
\end{aligned}
\label{eq:context-inheritance}
\end{equation}
The inner $\Compose$ replaces child $j$ in the parent snapshot with
$Q$; the outer context places the updated parent in its inherited
surroundings. At the root, the candidate program already describes
the entire world. At deeper levels, the same rule assembles the
candidate through the program snapshots along its ancestor path.
Thus $\mathcal C_j^{(k+1)}$ is fully determined by those snapshots
and the child references.

During a call, the surrounding snapshots remain fixed while the
candidate component is edited. These snapshots and the assembly
function define the construction-time inspection environment.
The component code, surrounding structures, and shared dependencies
are retained through the world program's references.

\subsubsection{Prepare and recursively construct the parts}
\label{sec:descent}
\label{sec:recurrence}

Lines~\ref{algline:select}--\ref{algline:recurse} of
Algorithm~\ref{alg:rcwm} select unresolved children
$\mathcal U_i^{(k)}\subseteq\operatorname{Children}(i)$ and create
or reopen their programs. Let $\mathcal G^{-}$ denote the preceding
whole-construction pass together with child preparation, and let
an overbar $\bar ~$ mark the updated parent state and its fields. This
phase produces the parent state and the prepared child
state--context pairs:
\begin{equation}
\left(
\bar q_i^{(k)},\;
\{(q_j^{(k+1)},\mathcal C_j^{(k+1)})\}_{j\in\mathcal U_i^{(k)}}
\right)
=
\mathcal G^{-}\!\left(q_i^{(k)},\mathcal C_i^{(k)};A\right).
\label{eq:whole-before}
\end{equation}

\paragraph{Matched reference and camera crops.}
Each child receives a reference crop and a camera view of the
corresponding scene region. Let $W=(x,y,w,h)$ denote a window
with upper-left coordinate $(x,y)$, width $w$, and height $h$,
and let $s>0$ be its magnification. For an input image $I$ and
output coordinates $(\alpha,\beta)$, the image crop is
\begin{equation}
\bigl[\operatorname{Crop}_{W,s}(I)\bigr](\alpha,\beta)
=
I\!\left(x+\frac{\alpha}{s},\,y+\frac{\beta}{s}\right),
\label{eq:image-crop}
\end{equation}
with output size $sw\times sh$.

\begin{figure}[t]
\centering
\includegraphics[width=\linewidth]{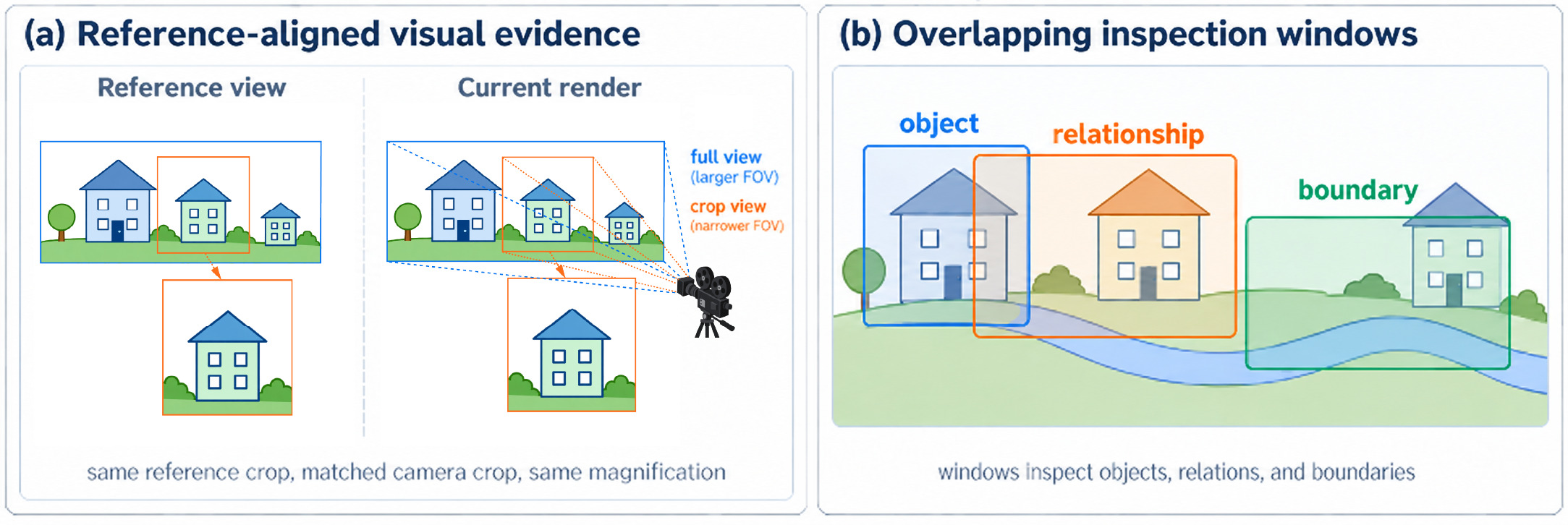}
\caption{\textbf{Visual inspection for recursive world construction.}
(a) Reference and camera crops use matching coordinates and
magnification, with the camera pose inherited from the parent.
(b) Overlapping windows inspect object details, spatial relationships,
and contact boundaries. Both panels are schematic.}
\label{fig:observations}
\end{figure}

For a perspective camera $\kappa=(R,t,K,(n_x,n_y))$, where
$R,t$ specify its world-to-camera pose, $K$ its intrinsic matrix,
and $(n_x,n_y)$ its viewport size, define
\begin{align}
H_{W,s}
&=
\begin{bmatrix}
s&0&-sx\\
0&s&-sy\\
0&0&1
\end{bmatrix},
\label{eq:crop-transform}\\
\operatorname{CropCamera}_{W,s}(\kappa)
&=
\left(R,t,H_{W,s}K,(sw,sh)\right).
\label{eq:camera-crop}
\end{align}
These paired operations preserve the camera pose and match the
reference crop's coordinates and magnification. Orthographic cameras
use the corresponding projection subwindow. Windows can overlap
to inspect objects, relationships, and contact boundaries
(Figure~\ref{fig:observations}).

\paragraph{Child inputs and recursive calls.}
For each selected child $j$, the agent chooses a window
$W_j^{(k+1)}$, magnification $s_j^{(k+1)}$, and budget
$b_j^{(k+1)}$, reserving resources for parent revisitation.
Its context $\mathcal C_j^{(k+1)}$ is derived from the updated
parent program through Equation~\ref{eq:context-inheritance}.
The child starts from the referenced component code and matched
visual evidence:
\begin{equation}
\begin{aligned}
P_j^{(k+1)}
&=\bar P_i^{(k)}[j],\\
I_j^{(k+1)}
&=\operatorname{Crop}_{W_j^{(k+1)},s_j^{(k+1)}}
  \!\left(I_i^{(k)}\right),\\
\kappa_j^{(k+1)}
&=\operatorname{CropCamera}_{W_j^{(k+1)},s_j^{(k+1)}}
  \!\left(\bar\kappa_i^{(k)}\right),\\
q_j^{(k+1)}
&=\left(P_j^{(k+1)},I_j^{(k+1)},
        \kappa_j^{(k+1)},b_j^{(k+1)}\right).
\end{aligned}
\label{eq:child-state}
\end{equation}
The parent then calls the same complete solver
(Algorithm~\ref{alg:rcwm}, line~\ref{algline:recurse}):
\begin{equation}
q_j^{(k+1)+}
=
\mathcal F(q_j^{(k+1)},\mathcal C_j^{(k+1)};A),
\qquad j\in\mathcal U_i^{(k)}.
\label{eq:recursive-call}
\end{equation}
Each child establishes its own whole, recursively solves finer
parts where needed, and refines their composition before returning.
Visual evidence guides further descent, allowing the same
construction process to continue at depth $k+2$ and beyond.

\subsubsection{Compose, refine, and return}
\label{sec:revisit}
\label{sec:reconcile}

Lines~\ref{algline:compose}--\ref{algline:visual-return} of
Algorithm~\ref{alg:rcwm} combine the returned child programs and
refine their relationships at the parent's depth $k$.

\paragraph{Compose the returned programs.}
Let $P_j^{(k+1)+}$ be the program returned by child $j$.
A tilde marks the assembled parent before refinement, and
$\widetilde b_i^{(k)}$ is the remaining budget after deducting
all child-call usage, including descendants, from $\bar b_i^{(k)}$.
The assembled program and state are
\begin{equation}
\begin{aligned}
\widetilde P_i^{(k)}
&=\Compose\!\left(
  \bar P_i^{(k)},
  \{j\mapsto P_j^{(k+1)+}\}_{j\in\mathcal U_i^{(k)}}
\right),\\
\widetilde q_i^{(k)}
&=\left(\widetilde P_i^{(k)},I_i^{(k)},
        \bar\kappa_i^{(k)},\widetilde b_i^{(k)}\right).
\end{aligned}
\label{eq:composition}
\end{equation}

\paragraph{Joint visual refinement.}
Let $\mathcal G^{+}$ denote the parent's inspection-and-editing
phase and $\sigma_i^{(k)}\in\{\textsc{continue},\textsc{return}\}$
its visual review decision:
\begin{equation}
\left(q_i^{(k)+},\sigma_i^{(k)}\right)
=
\mathcal G^{+}\!\left(
  \widetilde q_i^{(k)},\mathcal C_i^{(k)};A
\right).
\label{eq:whole-after}
\end{equation}
Using $\Observe(\widetilde q_i^{(k)},\mathcal C_i^{(k)})$,
the agent compares the assembled scene with its reference and
revises geometry, placement, and shared structures. For example,
a refined sign may need repositioning to avoid covering a
neighboring window. Comparisons are refreshed after edits.

The review checks both whether reference structures are correctly
represented and whether generated elements have reference support
or an explicit completion assumption. Shared structures are repaired
by their owning ancestor, and camera corrections are handled at
the root. Affected child contexts and views are refreshed before
further calls.

\paragraph{Continue or return.}
A complete review with no new actionable discrepancy produces
$\sigma_i^{(k)}=\textsc{return}$. Otherwise, another cycle starts
from $q_i^{(k)}\gets q_i^{(k)+}$ while budget remains.
Budget exhaustion returns the latest executable state.
The final root state $q_0^{(0)+}$ supplies the outputs
$P=P_0^{(0)+}$ and $\kappa=\kappa_0^{(0)+}$
(Algorithm~\ref{alg:rcwm}, line~\ref{algline:deliver}).

\section{Related Work}
\label{sec:related}

\paragraph{Code-based world representation and generation.}
Code represents worlds as executable, editable, and compositional programs rather than only rendered appearances. By exposing geometry, materials, and spatial relationships as explicit program elements, it supports targeted editing, component reuse, and integration with downstream simulation \citep{yin2026viga}. Code World Model further couples code-maintained world state with visually generated observations, separating executable world evolution from visual realization \citep{chen2026codeworld}. Infinigen demonstrates the richness of procedural world construction through compositional geometry and materials \citep{raistrick2023infinigen}. For text-driven generation, SceneCraft produces Blender programs through planning, visual refinement, and reusable code \citep{hu2024scenecraft}, while WorldClaw constructs open worlds through global planning, terrain construction, local content generation, and render-based refinement \citep{guo2026worldclaw}. These approaches demonstrate how code can support the construction of structured worlds that remain accessible beyond rendering. Unlike open-ended generation from procedural or textual specifications, our setting seeks an executable scene program whose rendering faithfully matches a particular reference image.

\paragraph{Image-to-code world reconstruction.}
Reconstructing a world as code from an image requires more than semantic plausibility: the program must reproduce specific visible shapes, spatial arrangements, occlusions, and fine details under a consistent camera, despite ambiguity in the underlying 3D geometry \citep{yin2026viga,he2026thinking}. Recent vision-language coding agents address this problem through execution-grounded visual feedback. VIGA reconstructs and edits scenes through interleaved program generation, execution, and visual verification \citep{yin2026viga}. Thinking in Blender combines scene-graph initialization with factor-specific reconstruction stages for geometry, materials, composition, and lighting \citep{he2026thinking}. The img2threejs project provides a reference-driven, code-only workflow for procedural object reconstruction in Three.js \citep{img2threejs}. \method{} constructs \emph{Recursive Scene Programs} through a recursive solver: unresolved subworlds receive the same complete solve, and each return triggers renewed inspection and refinement of the assembled parent scene. The recurrence therefore encompasses the entire construction process, rather than only scene decomposition, repeated program editing, or a fixed sequence of reconstruction factors.

\paragraph{Recursive inference and concurrent work.}
Recursive Language Models organize inference through programmatic examination of external context and recursive model calls \citep{zhang2025rlm}. In our setting, the external environment is an evolving executable world: child calls construct subprograms whose return changes the scene that their parent must inspect. Concurrent work, FuncRoom-Agent, introduces a recursive DSL and distills construction traces into an expert specialized for indoor scene generation \citep{feng2026funcroom}. Its representation, construction stages, and training objectives are tailored to room structure, functional furniture arrangements, and nested indoor objects. In contrast, \method{} formulates recursion as a general world-construction principle rather than a room-specific generation procedure. A frozen coding model applies the same complete visual solver to objects, interacting assemblies, and continuous terrain, spanning architectural details, buildings, towns, and terrain-rich worlds without prescribing a domain-specific semantic hierarchy. Recursion remains an active inference-time process: each subworld can invoke further solves and refine its assembled whole before returning to its parent.
% Experiments: reconstruction results first; construction organization in ablations.
% Rewritten from the attached September 10, 08:14 text.
% Numerical table entries and figure asset paths are unchanged.
% sections/variants-table remains an existing project dependency.

\section{Experiments}
\label{sec:experiments}

We evaluate how faithfully \method{} reconstructs complex worlds as
executable scene programs, considering scene-wide appearance and
fine-scale detail. We first compare its reconstructions with
image-to-scene-program baselines (Section~\ref{sec:casestudy}), then
examine the roles of initial whole-scene construction, recursive
child solves, and parent revisitation through ablations
(Section~\ref{sec:controls}). Additional views and recorded call trees
illustrate the generated geometry and the construction process.

\subsection{Setup}
\label{sec:setup}

\paragraph{Reference images.}
We use five whole-scene references and five local crops. The
\emph{city-full} reference and four crops---\emph{school-block},
\emph{police-corner}, \emph{park-lake}, and \emph{shop-row}---come
from the example composition of the CC0 ``Isometric city'' sprite pack
by JanaChumi \citep{janachumi2017isometric}. The other references are
WorldClaw demonstration renders \citep{guo2026worldclaw}:
\emph{island-harbor} (Fig.~4), \emph{medieval-village} (Fig.~9),
\emph{snow-village} (Fig.~10), \emph{japan-island} (Fig.~12), and the
\emph{valley-village} crop from Fig.~15. These supply reference images
only; reconstruction uses the images without their source prompts,
terrain, or assets. All methods receive the same raw images at the
same resolution. View tabs, biome badges, and location pins remain
in the inputs and evaluation references.

\paragraph{Methods and execution.}
We compare \method{} with SEIG \citep{he2026thinking}, VIGA
\citep{yin2026viga}, and img2threejs \citep{img2threejs}, using the same
base model and reasoning-effort setting: every method is driven by
OpenAI's \texttt{gpt-6-astra} through the Codex command-line agent at
high reasoning effort, and \method{} runs each node of the recursion as
one Codex session. \method{} uses the instruction in
Appendix~\ref{app:solver}.  SEIG is our reproduction from the paper. VIGA and img2threejs use
their released pipelines with the shared model endpoint. %VIGA's agents build assets procedurally in Blender, with its commercial asset-generation service disabled. Under its native stopping rules, img2threejs completes only \emph{city-full}; its reference-suitability gate or blockout-correction limit stops the other nine cases. For those cases, we resume the same session with one instruction to continue and deliver a render, without further assistance, and report the resulting outputs.

\paragraph{Evaluation.} We report PSNR, SSIM, edge $F_1$ using Canny edges with a three-pixel tolerance,
LPIPS with the AlexNet backbone \citep{zhang2018lpips}, and cosine
similarity of CLIP ViT-B/32 image embeddings. Metrics are computed
against the unmodified reference at its native resolution; best values
per scene are shown in bold. Visual comparisons show full views and
magnified details at matched scales. The ablation additionally reports
mean SSIM over eight fixed $2\times$ detail windows. Spatial
relationships and program editability are inspected qualitatively in
the renders and source programs. The solver uses visual feedback
during construction; these metrics are computed for evaluation.

\subsection{Reconstruction results}
\label{sec:casestudy}

\paragraph{Reference-view fidelity.} Table~\ref{tab:casemetrics} compares all four methods on the whole scenes and local crops. In these runs, \method{} obtains the highest PSNR and lowest LPIPS on every reference. It also has the highest SSIM on nine references; on \emph{valley-village}, VIGA obtains 0.41 and \method{} obtains 0.38. The PSNR and LPIPS improvements span both the whole-scene and
local-crop groups.

\begin{figure}[t]
\centering
\includegraphics[width=\linewidth]{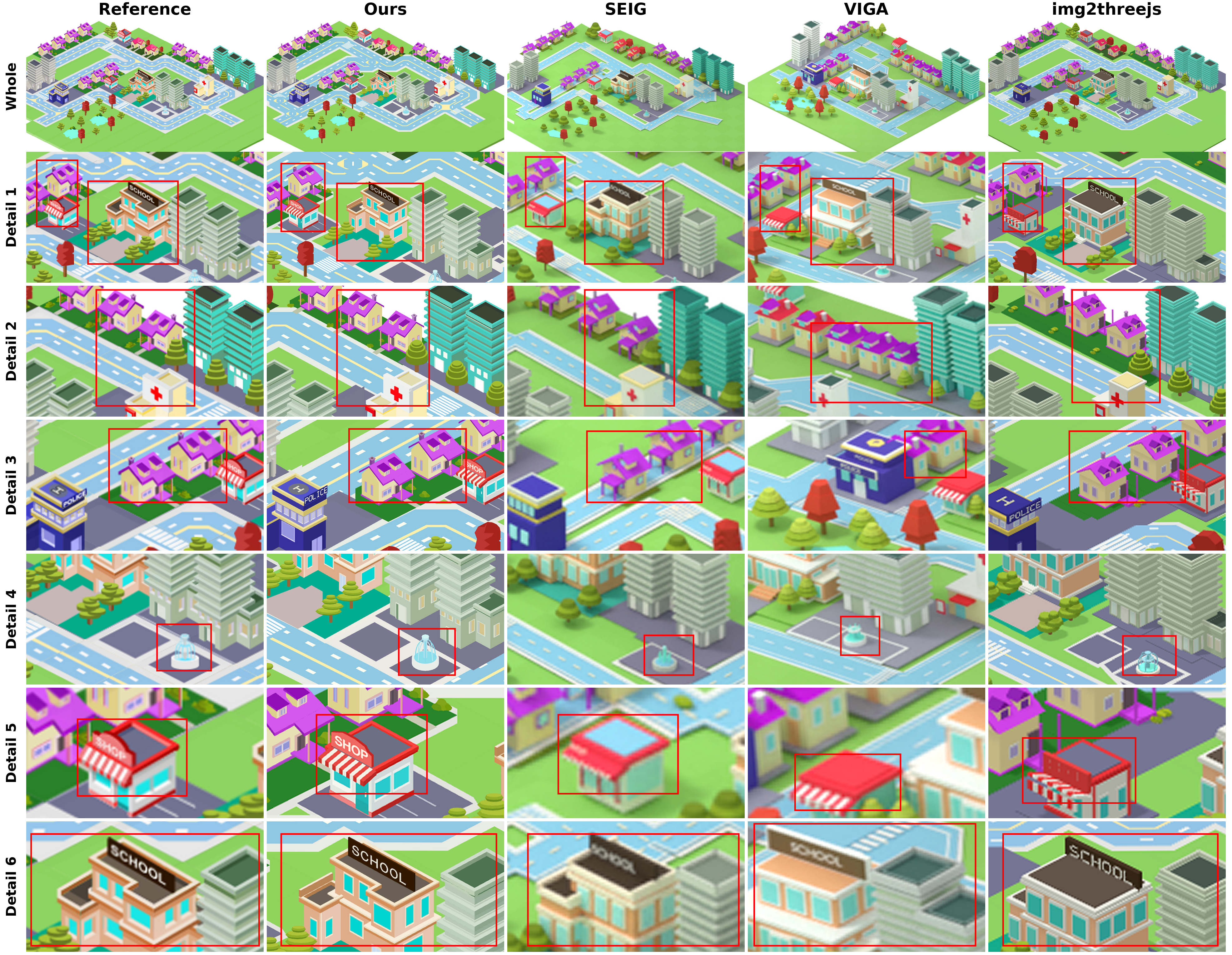}
\caption{Whole-city case: full views (top) and six affine-aligned magnified windows from medium to small (below), all methods on the same base model and raw input. Reference: example composition of the CC0 ``Isometric city'' sprite pack \citep{janachumi2017isometric}.}
\label{fig:case-city-full}
\end{figure}

\begin{figure}[t]
\centering
\includegraphics[width=\linewidth]{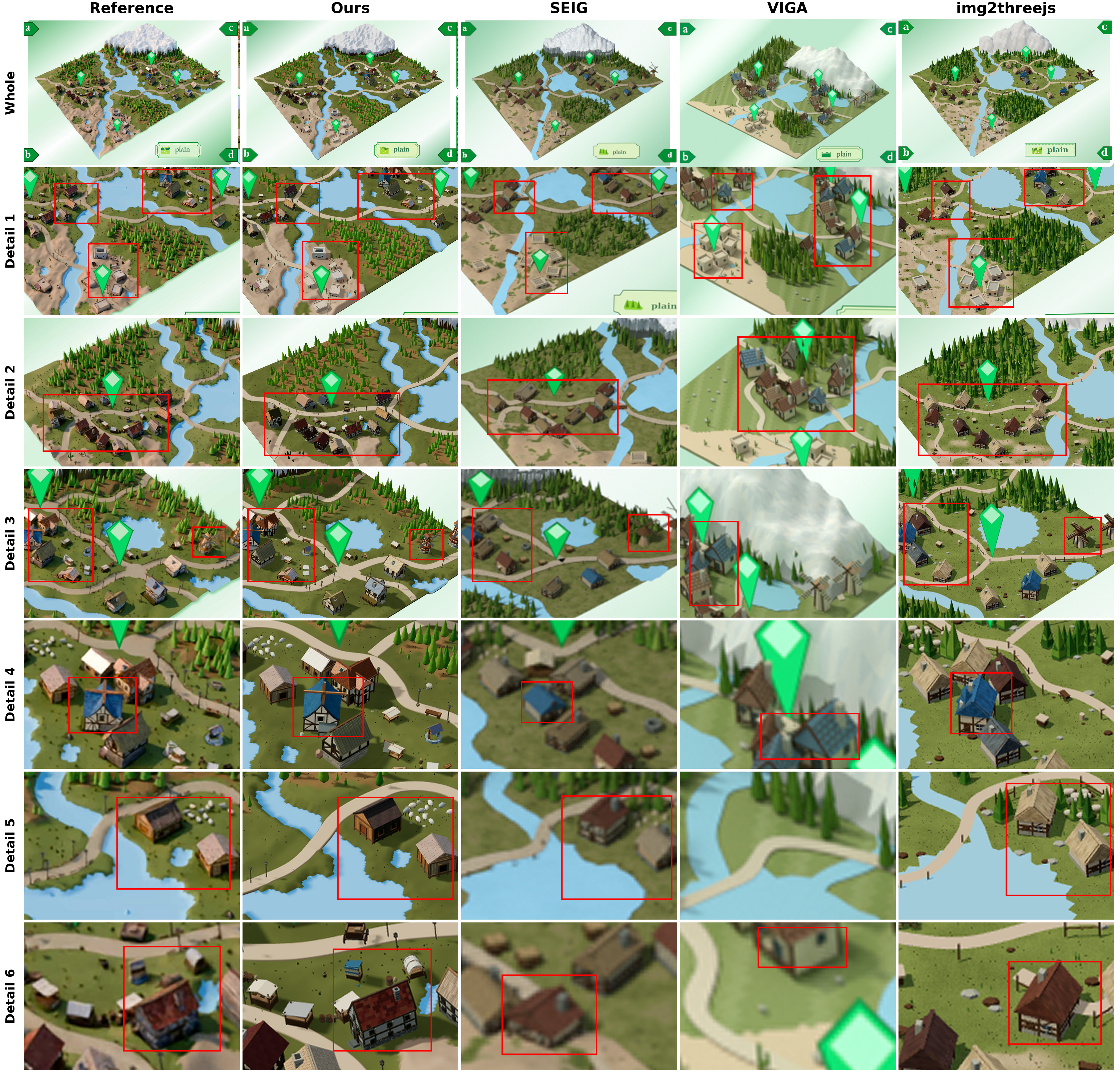}
\caption{Medieval-village case: full views (top) and six affine-aligned magnified windows from medium to small (below), all methods on the same base model and raw input. Reference: WorldClaw Fig.~9 \citep{guo2026worldclaw}.}
\label{fig:case-medieval-village}
\end{figure}

\begin{figure}[t]
\centering
\vspace{-14mm}
\includegraphics[width=0.45\linewidth]{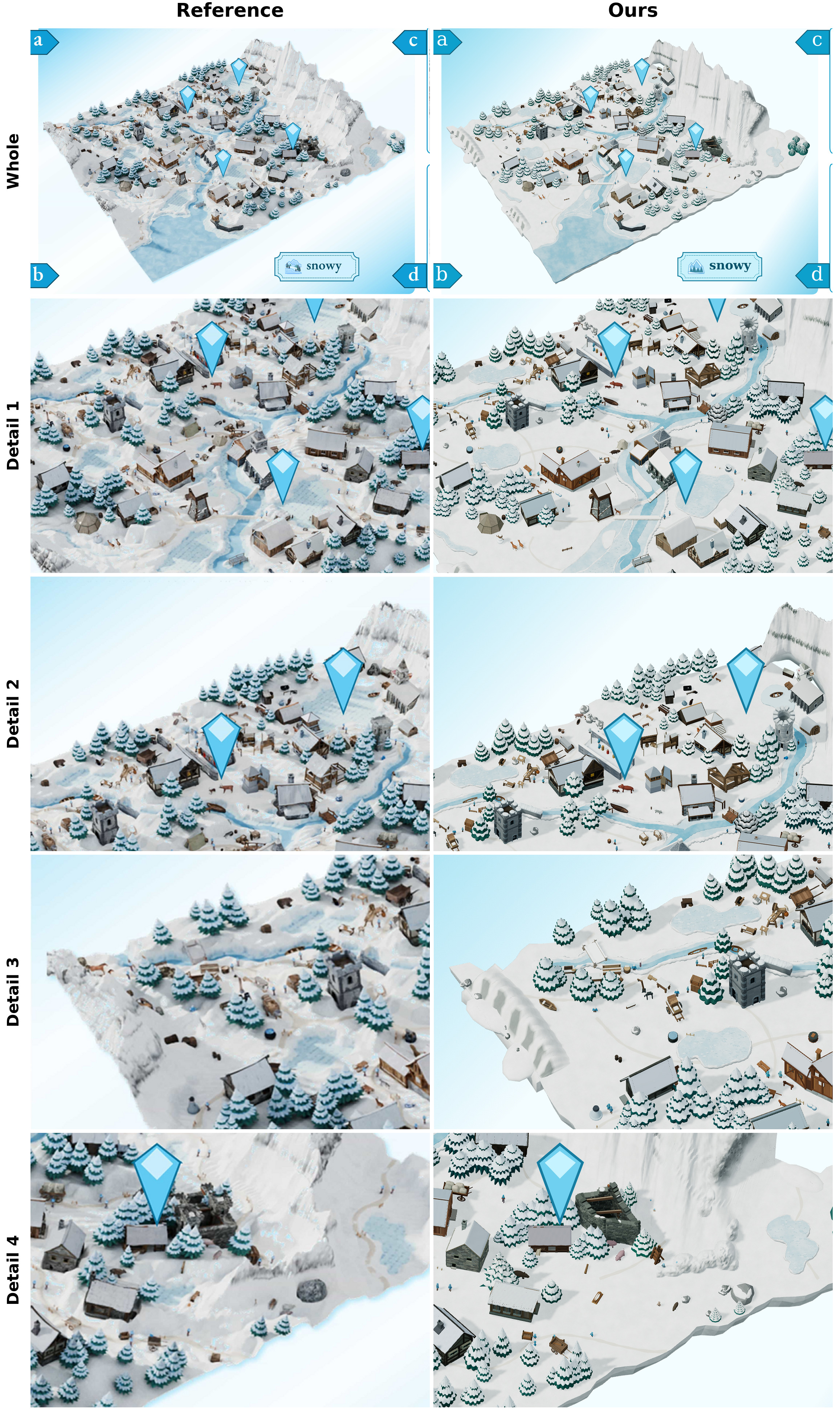}
\includegraphics[width=0.45\linewidth]{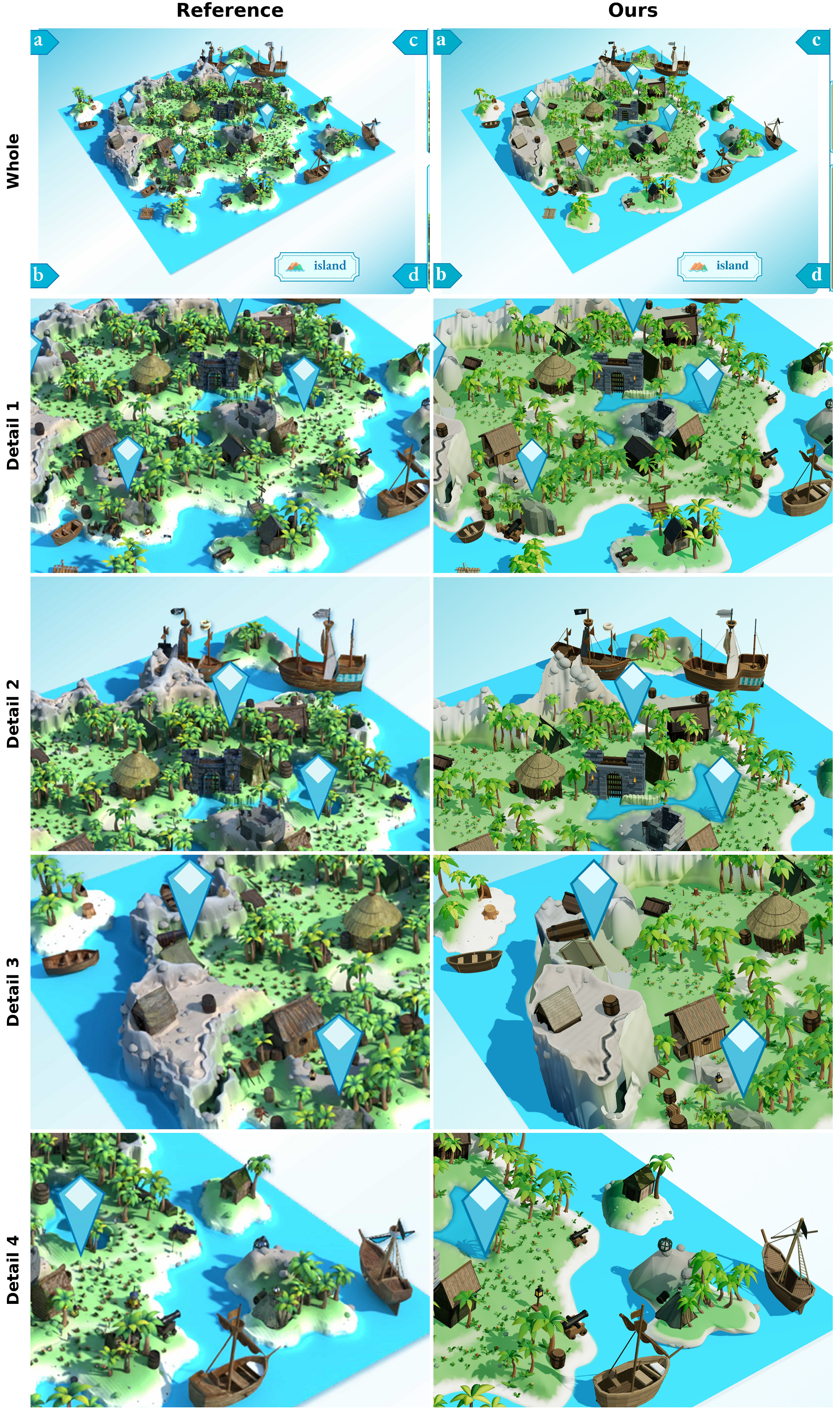}
\caption{Reference and our result, full view on top and four magnified windows side by side: \emph{snow-village} (left; WorldClaw Fig.~10) and \emph{island-harbor} (right; WorldClaw Fig.~4) \citep{guo2026worldclaw}.}
\label{fig:ours-a}
\end{figure}

\begin{figure}[t]
\centering
\vspace{-14mm}
\includegraphics[width=0.45\linewidth]{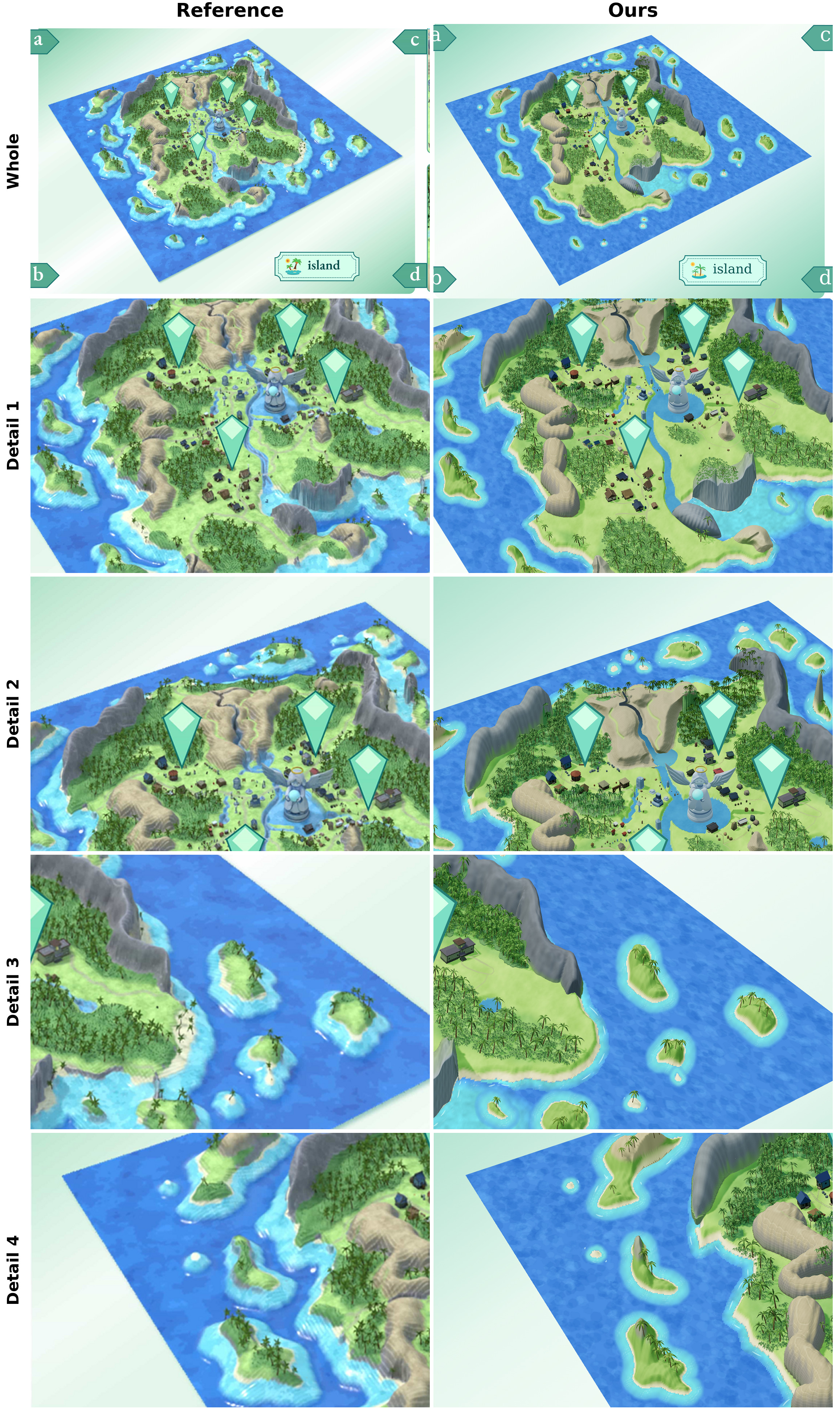}
\includegraphics[width=0.45\linewidth]{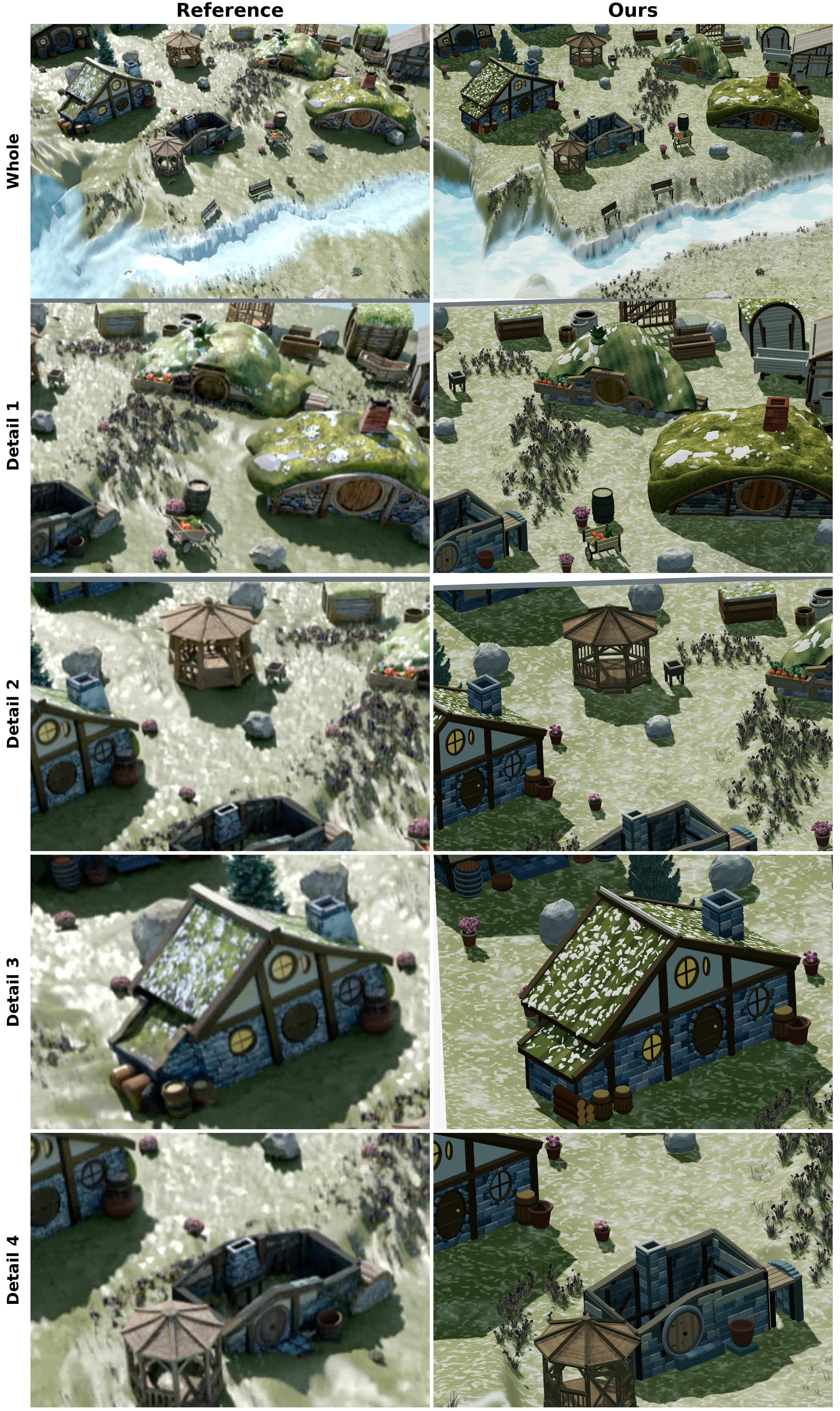}
\caption{Reference and our result, full view on top and four magnified windows side by side: \emph{japan-island} (left; WorldClaw Fig.~12) and \emph{valley-village} (right; a crop of WorldClaw Fig.~15) \citep{guo2026worldclaw}.}
\label{fig:ours-b}
\end{figure}

\begin{figure}[t]
\centering
\includegraphics[width=0.45\linewidth]{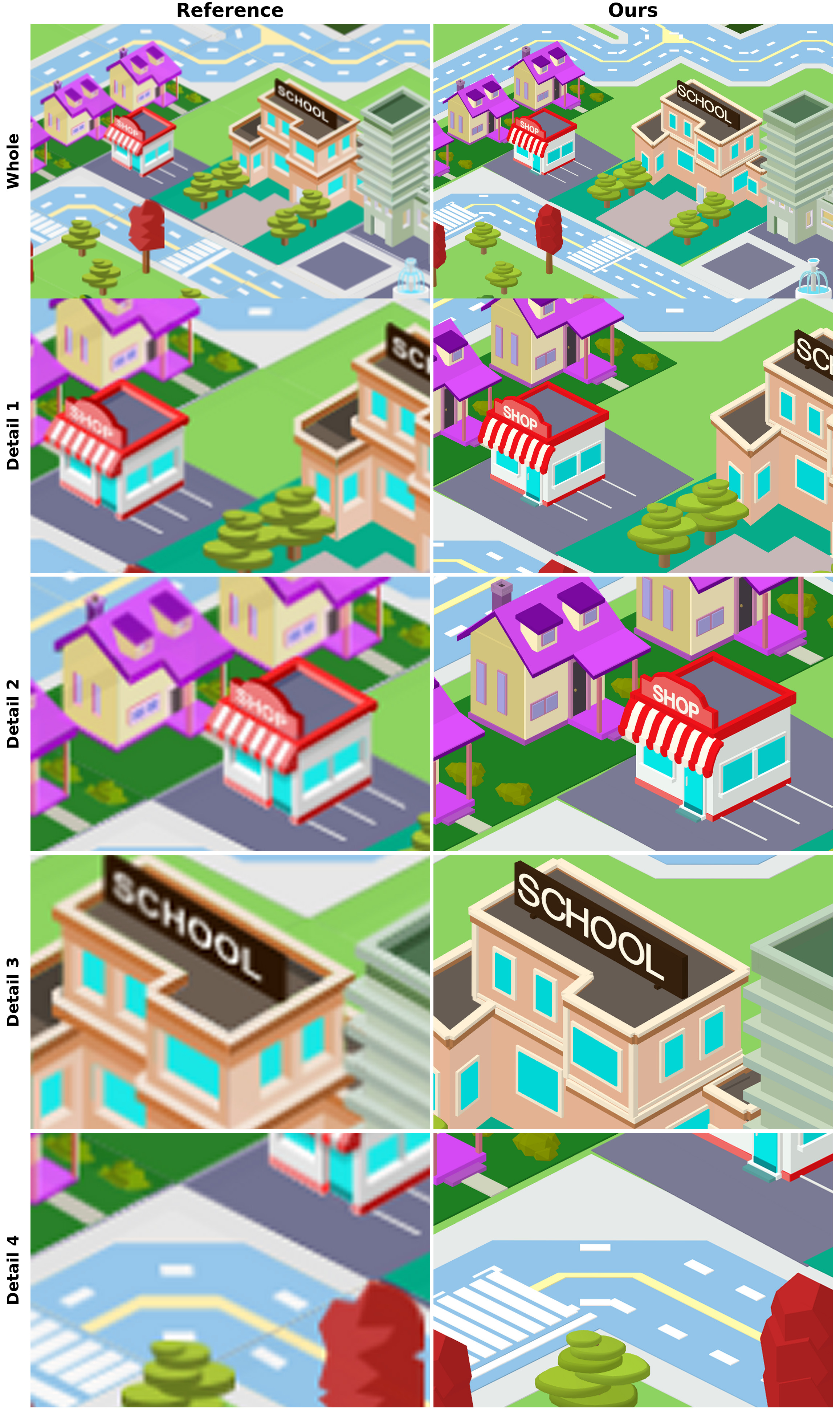}\hfill
\includegraphics[width=0.45\linewidth]{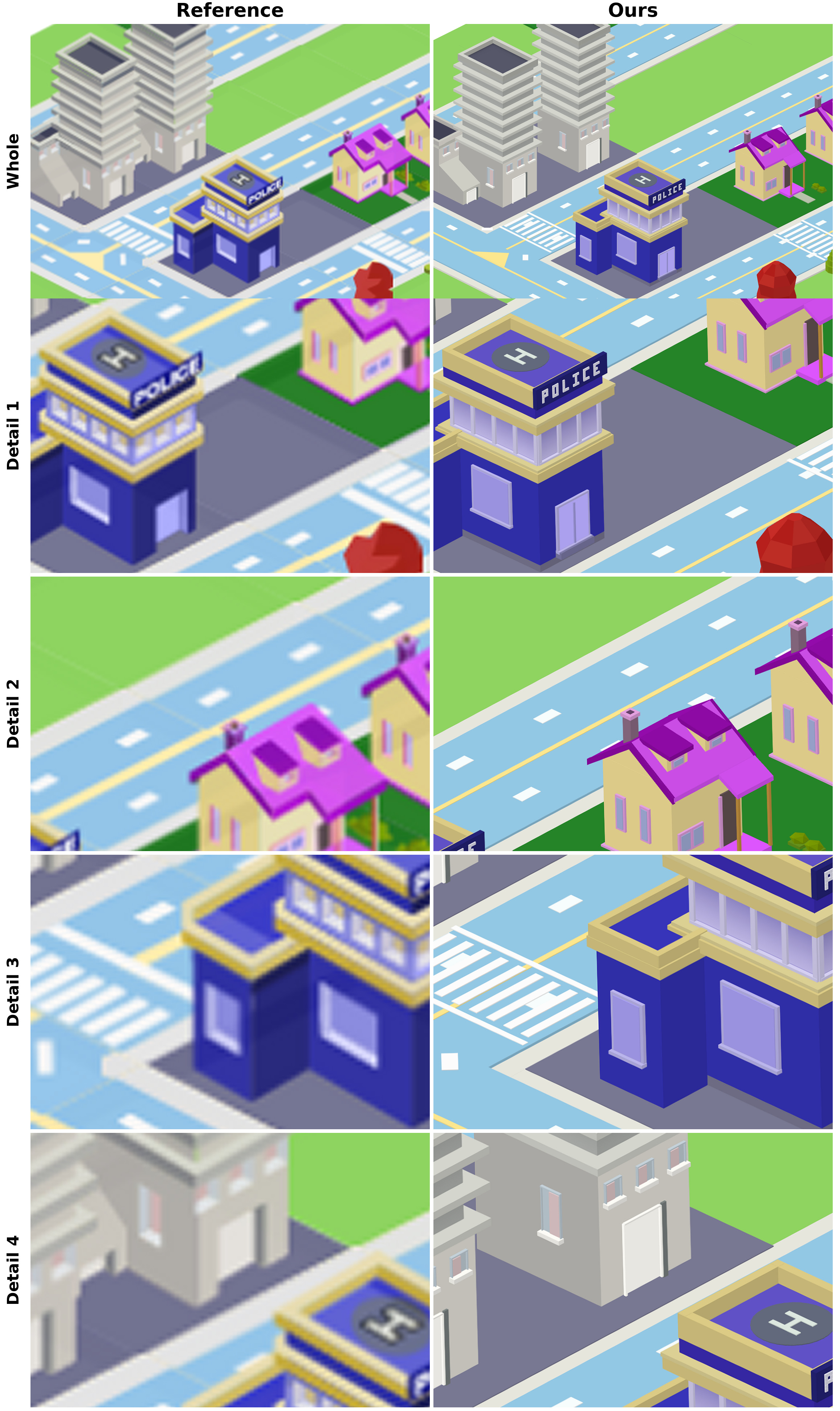}
\caption{Reference and our result, full view on top and four magnified windows side by side: \emph{school-block} (left) and \emph{police-corner} (right), crops of the ``Isometric city'' sprite-pack example \citep{janachumi2017isometric}.}
\vspace{-4mm}
\label{fig:ours-c}
\end{figure}

\begin{figure}[t]
\centering
\includegraphics[width=0.45\linewidth]{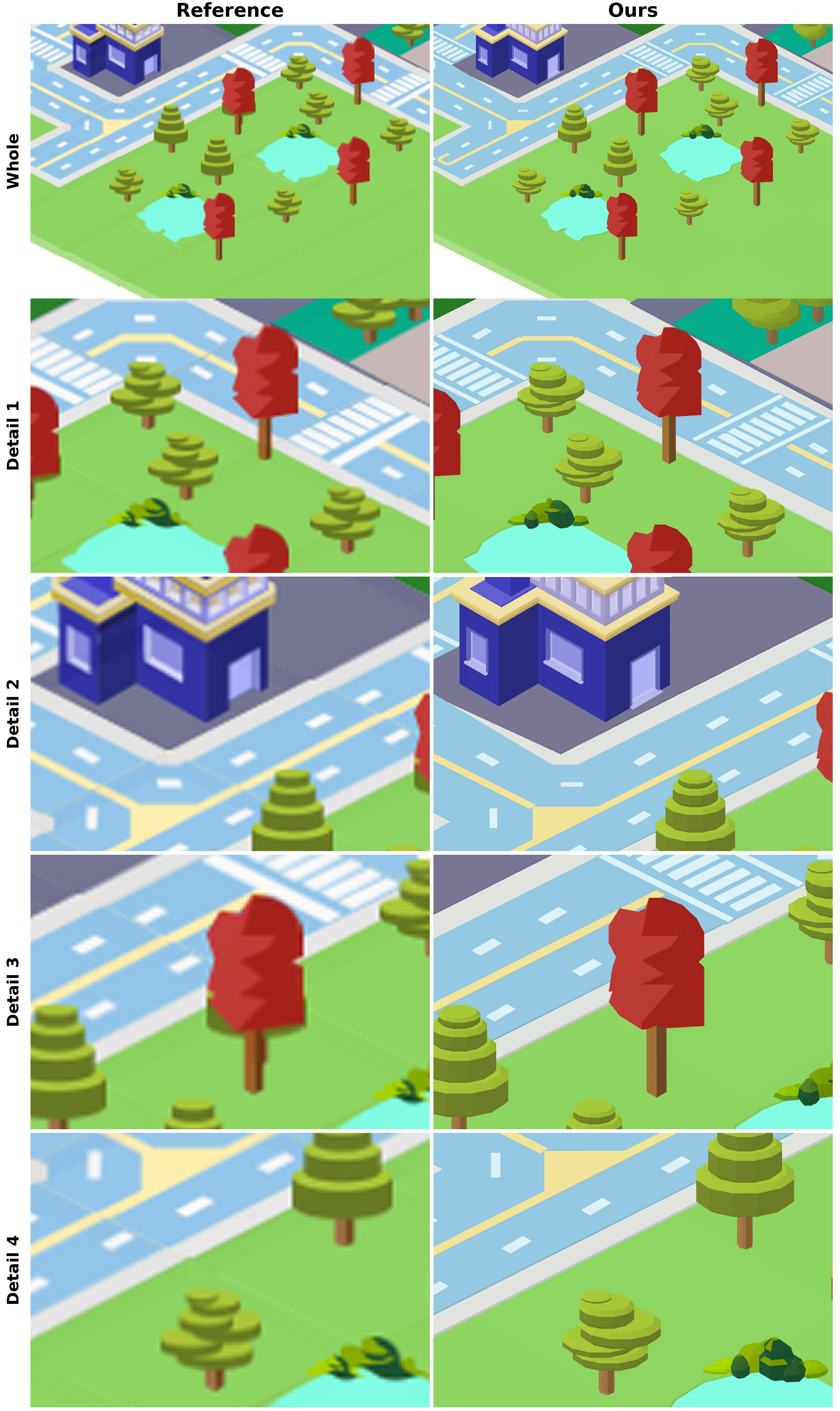}\hfill
\includegraphics[width=0.45\linewidth]{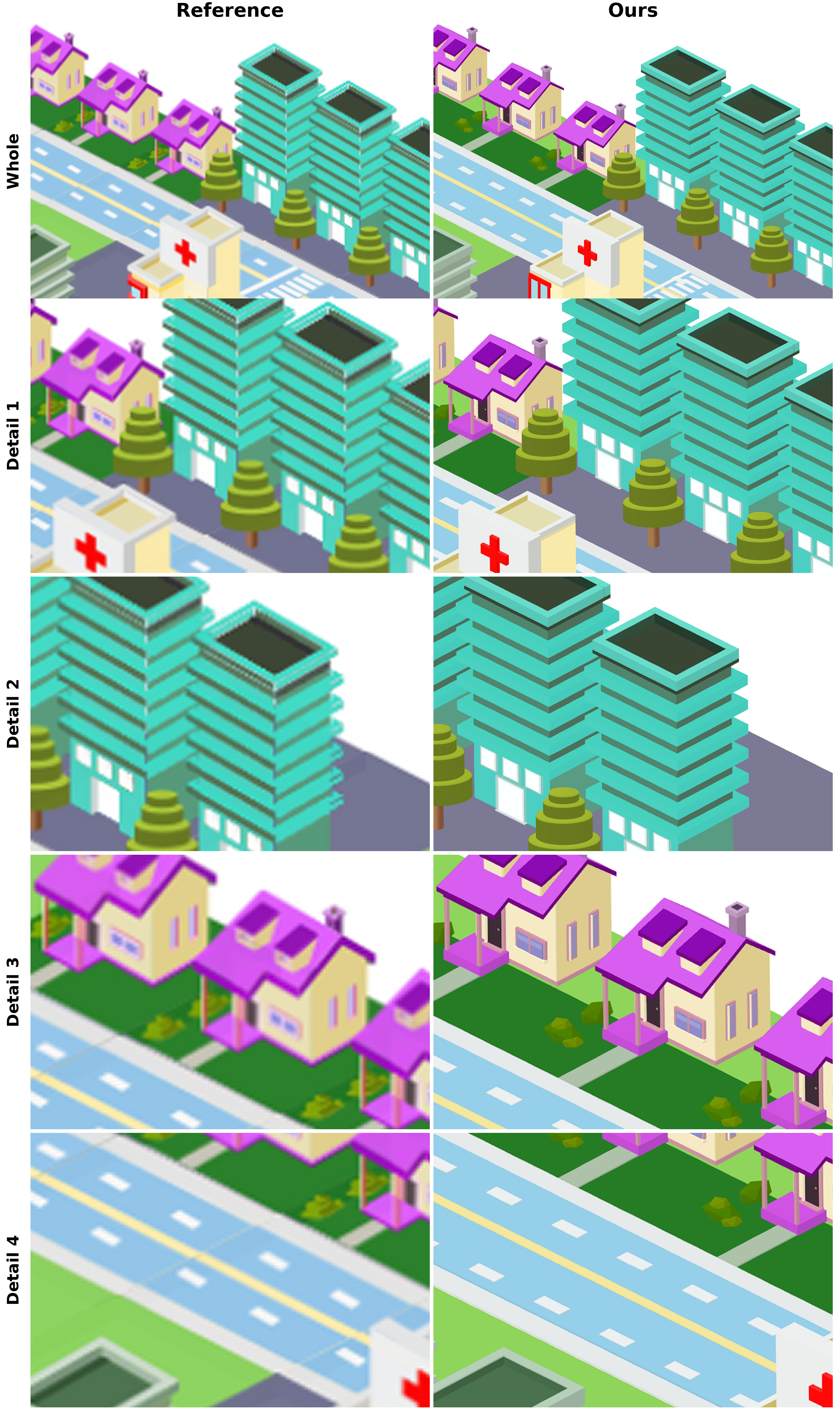}
\caption{Reference and our result, full view on top and four magnified windows side by side: \emph{park-lake} (left) and \emph{shop-row} (right), crops of the ``Isometric city'' sprite-pack example \citep{janachumi2017isometric}.}
\label{fig:ours-d}
\end{figure}

\paragraph{Whole scenes and local detail.}
Figures~\ref{fig:case-city-full} and~\ref{fig:case-medieval-village}
show all methods on \emph{city-full} and \emph{medieval-village}.
Full views show the overall reconstruction, followed by six
affine-aligned magnified windows selected to highlight differences
between methods. Figures~\ref{fig:ours-a}--\ref{fig:ours-d} pair the
remaining eight references with our results and four magnified windows
per case. Together, these views allow scene layout and smaller
structures, including individual buildings, groves, and shorelines,
to be inspected at their respective scales. 

\begin{table}[t]
\caption{Reconstruction metrics for every method with a completed run; best per scene in bold.}
\label{tab:casemetrics}
\small
\begin{tabularx}{\linewidth}{@{}l l Y Y Y Y Y@{}}
\toprule
Scene & Method & PSNR$\uparrow$ & SSIM$\uparrow$ & Edge $F_1\uparrow$ & LPIPS$\downarrow$ & CLIP$\uparrow$ \\
\midrule
\multicolumn{7}{@{}l}{\emph{Whole scenes}}\\
city-full & Ours & \textbf{18.2} & \textbf{0.66} & \textbf{0.94} & \textbf{0.158} & \textbf{0.95} \\
 & SEIG & 14.5 & 0.51 & 0.83 & 0.339 & 0.90 \\
 & VIGA & 9.5 & 0.48 & 0.58 & 0.670 & 0.83 \\
 & img2threejs & 15.5 & 0.55 & 0.91 & 0.243 & 0.93 \\
\midrule
snow-village & Ours & \textbf{17.9} & \textbf{0.73} & \textbf{0.85} & \textbf{0.237} & \textbf{0.83} \\
 & SEIG & 13.4 & 0.64 & 0.37 & 0.520 & 0.77 \\
 & VIGA & 13.5 & 0.61 & 0.40 & 0.541 & 0.75 \\
 & img2threejs & 15.4 & 0.64 & 0.65 & 0.351 & 0.81 \\
\midrule
island-harbor & Ours & \textbf{17.4} & \textbf{0.74} & \textbf{0.90} & \textbf{0.186} & \textbf{0.93} \\
 & SEIG & 11.7 & 0.63 & 0.57 & 0.516 & 0.80 \\
 & VIGA & 10.6 & 0.65 & 0.48 & 0.526 & 0.83 \\
 & img2threejs & 15.0 & 0.68 & 0.81 & 0.272 & 0.91 \\
\midrule
medieval-village & Ours & \textbf{18.7} & \textbf{0.71} & \textbf{0.89} & \textbf{0.200} & \textbf{0.93} \\
 & SEIG & 10.9 & 0.60 & 0.50 & 0.541 & 0.85 \\
 & VIGA & 10.5 & 0.58 & 0.37 & 0.601 & 0.83 \\
 & img2threejs & 15.4 & 0.62 & 0.77 & 0.325 & 0.89 \\
\midrule
japan-island & Ours & \textbf{19.3} & \textbf{0.73} & \textbf{0.87} & \textbf{0.201} & \textbf{0.96} \\
 & SEIG & 13.3 & 0.64 & 0.47 & 0.513 & 0.93 \\
 & VIGA & 12.9 & 0.65 & 0.43 & 0.534 & 0.80 \\
 & img2threejs & 16.6 & 0.66 & 0.76 & 0.302 & 0.95 \\
\midrule
\multicolumn{7}{@{}l}{\emph{Local crops}}\\
school-block & Ours & \textbf{16.4} & \textbf{0.56} & \textbf{0.97} & \textbf{0.171} & \textbf{0.94} \\
 & SEIG & 11.8 & 0.23 & 0.80 & 0.448 & 0.88 \\
 & VIGA & 11.2 & 0.19 & 0.77 & 0.587 & 0.85 \\
 & img2threejs & 12.8 & 0.26 & 0.88 & 0.310 & 0.91 \\
\midrule
police-corner & Ours & \textbf{16.6} & \textbf{0.44} & \textbf{0.97} & \textbf{0.159} & \textbf{0.91} \\
 & SEIG & 12.7 & 0.24 & 0.75 & 0.415 & 0.90 \\
 & VIGA & 12.4 & 0.27 & 0.82 & 0.473 & 0.91 \\
 & img2threejs & 13.7 & 0.30 & 0.86 & 0.285 & 0.90 \\
\midrule
park-lake & Ours & \textbf{23.3} & \textbf{0.83} & \textbf{0.99} & \textbf{0.075} & \textbf{0.95} \\
 & SEIG & 12.3 & 0.48 & 0.61 & 0.490 & 0.90 \\
 & VIGA & 15.1 & 0.54 & 0.74 & 0.330 & 0.92 \\
 & img2threejs & 18.4 & 0.66 & 0.92 & 0.169 & 0.93 \\
\midrule
shop-row & Ours & \textbf{17.4} & \textbf{0.65} & \textbf{0.96} & \textbf{0.120} & \textbf{0.96} \\
 & SEIG & 12.5 & 0.50 & 0.79 & 0.400 & 0.89 \\
 & VIGA & 15.2 & 0.54 & 0.91 & 0.249 & 0.91 \\
 & img2threejs & 16.9 & 0.59 & 0.94 & 0.145 & 0.95 \\
\midrule
valley-village & Ours & \textbf{14.1} & 0.38 & \textbf{0.48} & \textbf{0.583} & \textbf{0.87} \\
 & SEIG & 9.7 & 0.39 & 0.22 & 0.742 & 0.64 \\
 & VIGA & 10.2 & \textbf{0.41} & 0.20 & 0.743 & 0.68 \\
 & img2threejs & 11.3 & 0.34 & 0.41 & 0.726 & 0.70 \\
\bottomrule
\end{tabularx}
\end{table}

\begin{figure}[t]
\centering
\includegraphics[width=1\linewidth]{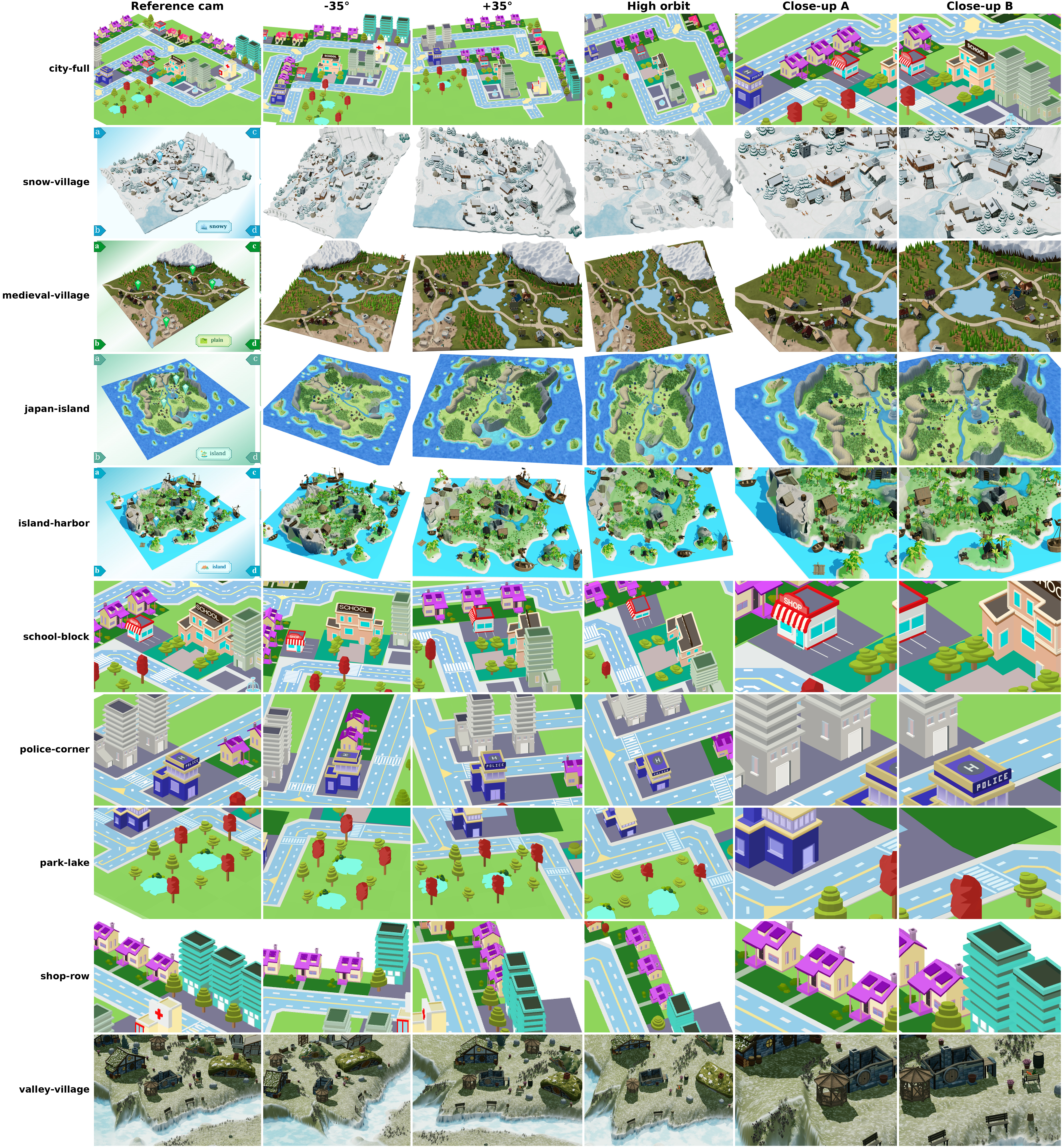}
\caption{Additional views of the delivered scene programs. Each row renders one program from the reference camera, two azimuth rotations of $35^\circ$, a raised orbit, and two close-ups, through one generic camera harness and without any re-optimization.Crop-based scenes only contain content visible in their reference, so unseen regions are absent when viewed from behind.} %The programs for the WorldClaw-derived scenes model the figure furniture (view tabs, biome badge) as a screen-space overlay layer: it is on in the reference-camera column and hidden in the other five views, whereas the location pins are anchored in world coordinates and move with the scene. 
\label{fig:novelviews}
\end{figure}

\begin{figure}[t]
\centering
\includegraphics[width=\linewidth]{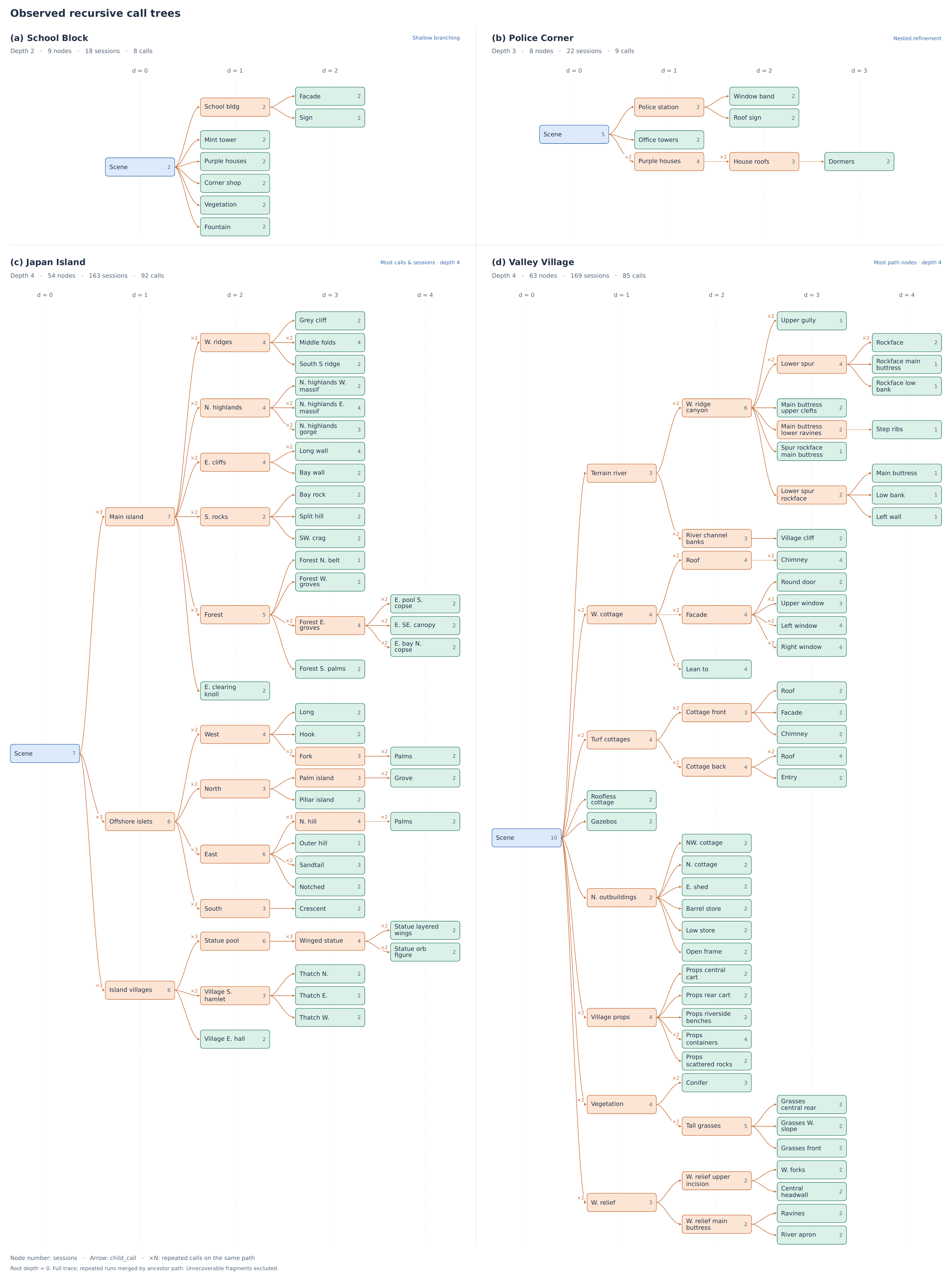}
\caption{Solver-call trees reconstructed from runner-logged call and return
 events for four representative cases. The logged internal nodes follow
 the complete whole--parts--whole cycle. Trees for all ten scenes are
 released with the code.}
\label{fig:trees}
\end{figure}

\paragraph{Additional views and construction traces.}
To demonstrate the 3D structure of the reconstructed worlds,
Figure~\ref{fig:novelviews} renders the delivered programs from
rotated, elevated, and close-up cameras without further optimization.
These views reveal the geometry and spatial arrangement encoded in
the same scene program beyond the reference projection. Figure~\ref{fig:trees} shows the recursive call trees recorded during construction, including nested calls to deeper subworlds and returns
to their parents. These traces demonstrate the recursive execution of the same solver across multiple levels.

\subsection{Ablation study}
\label{sec:controls}

\paragraph{Construction organization.}
We compare five reconstruction workflows that vary the order of
whole-scene and local construction, refinement after composition,
and the permitted depth of recursive calls. \emph{Flat + zoom} reconstructs the entire scene within one editing
process, using crop and zoom tools to inspect local details.
\emph{Local $\rightarrow$ global} first reconstructs components
independently and then assembles them into the whole scene.
\emph{Global $\rightarrow$ local} first establishes the scene layout,
then refines individual components, and finishes by assembling their
results without further editing of the combined scene.
\emph{Recursive, fixed 2-level} first establishes the whole, gives
selected components their own solver calls, and then renders and
refines the assembled scene. This cycle can repeat, with calls
restricted to the root and its direct children; each child completes
its work locally.
\emph{Recursive, free depth} uses the same solver and additionally
allows each child to call it on its own unresolved parts. These
deeper calls follow the same process: establish their whole,
construct their parts, and refine the composition before returning.

All variants use the same base model, scene representation, modeling
tools, reference resolution, and crop access, with variant-specific
instructions executed by the same runner. The two recursive variants
use the same solver and differ in their depth restriction.

\paragraph{Results.}
Table~\ref{tab:variants} compares the five variants on
\emph{school-block} and \emph{medieval-village}, with one run per
variant and scene under the shared ablation setting.
On \emph{school-block}, the free-depth run uses only two levels—the
root and its direct children—and achieves scores similar to the
fixed two-level variant.
On \emph{medieval-village}, it expands to three levels with 33 nodes
and achieves higher fidelity: compared with the fixed two-level
variant, whole-frame PSNR increases from 16.8 to 19.0 and local
SSIM from 0.52 to 0.60. It also outperforms the other three
workflows on both measures.
Together, these results suggest that deeper recursive solves can
improve both local detail and whole-scene fidelity, with the
gains varying across reconstruction tasks.

\begin{table}[h]
\caption{Ablation study on two scenes, single runs on the same base model and scene compiler. Global fidelity is whole-frame PSNR/SSIM/LPIPS against the reference; local fidelity is the mean SSIM over eight fixed $2\times$ detail windows. Best per scene in bold.}
\label{tab:variants}
\centering
\small
\begin{tabular}{@{}lcccc@{}}
\toprule
Variant & PSNR$\uparrow$ & SSIM$\uparrow$ & Local SSIM$\uparrow$ & LPIPS$\downarrow$ \\
\midrule
\multicolumn{5}{@{}l}{\emph{school-block}}\\
Flat + zoom & 13.3 & 0.28 & 0.27 & 0.294 \\
Local $\rightarrow$ global & 16.3 & 0.53 & 0.52 & 0.179 \\
Global $\rightarrow$ local & 13.4 & 0.28 & 0.28 & 0.288 \\
\textbf{Recursive, fixed 2-level} & \textbf{16.8} & \textbf{0.55} & \textbf{0.57} & \textbf{0.166} \\
\textbf{Recursive, free depth} & 16.5 & 0.52 & 0.52 & 0.169 \\
\midrule
\multicolumn{5}{@{}l}{\emph{medieval-village}}\\
Flat + zoom & 16.4 & 0.66 & 0.50 & 0.261 \\
Local $\rightarrow$ global & 14.3 & 0.64 & 0.52 & 0.402 \\
Global $\rightarrow$ local & 14.9 & 0.65 & 0.52 & 0.377 \\
\textbf{Recursive, fixed 2-level} & 16.8 & 0.67 & 0.52 & 0.252 \\
\textbf{Recursive, free depth} & \textbf{19.0} & \textbf{0.73} & \textbf{0.60} & \textbf{0.187} \\
\bottomrule
\end{tabular}
\end{table}

\section{Conclusion and Limitations}
\label{sec:conclusion}
\label{sec:discussion}

We introduced Recursive Code World Models and Recursive Scene Programs
for reconstructing complex, executable 3D worlds from a single
reference image. The framework applies the same complete construction
process to each selected subworld: establish the whole, recursively
reconstruct unresolved parts, and revisit and refine their composition.
Reference-aligned camera views support detailed local inspection,
while parent revisitation addresses the spatial relationships among
the returned components. \method{} therefore combines the compositional
\rsp{} representation with a recursive construction process that
addresses local detail and whole-scene consistency at every scale.

Several limitations remain. Single-view input leaves hidden geometry
and absolute scale ambiguous, and visual inspection can overlook
errors. Repeated refinement and interactions among components can
require substantial computation, and convergence is not guaranteed.
Our evaluation covers a limited set of scenes, with single runs on
one base model and ablations on two scenes. Larger benchmarks,
repeated runs, and human evaluation of spatial relationships would
provide a stronger assessment of reconstruction quality and reliability.
\label{lastmainpage}
\clearpage
\bibliography{references}

@misc{chen2026codeworld,
  title         = {Code World Model: Coding Agent as World Brain},
  author        = {Chen, Yiwen and Lin, Guosheng and Zhang, Chi},
  year          = {2026},
  eprint        = {2608.25927},
  archivePrefix = {arXiv},
  primaryClass  = {cs.CV},
  url           = {https://arxiv.org/abs/2608.25927}
}

@misc{yin2026viga,
  title         = {Vision-as-Inverse-Graphics Agent via Interleaved Multimodal Reasoning},
  author        = {Yin, Shaofeng and Ge, Jiaxin and Wang, Zora Zhiruo and Wang, Chenyang and Li, Xiuyu and Black, Michael J. and Darrell, Trevor and Kanazawa, Angjoo and Feng, Haiwen},
  year          = {2026},
  eprint        = {2601.11109},
  archivePrefix = {arXiv},
  primaryClass  = {cs.CV},
  note          = {Version 3, revised April 6, 2026},
  url           = {https://arxiv.org/abs/2601.11109}
}

@misc{he2026thinking,
  title         = {Thinking in {Blender}: Staged Executable Inverse Graphics with Vision-Language Models},
  author        = {He, Guangzhao and Luo, Rundong and Ma, Wei-Chiu and Averbuch-Elor, Hadar},
  year          = {2026},
  eprint        = {2606.02580},
  archivePrefix = {arXiv},
  primaryClass  = {cs.CV},
  url           = {https://arxiv.org/abs/2606.02580}
}

@misc{img2threejs,
  title         = {{img2threejs}},
  author        = {{img2threejs contributors}},
  year          = {2026},
  version       = {2.0.0},
  howpublished  = {GitHub repository},
  note          = {Accessed September 8, 2026},
  url           = {https://github.com/img2threejs/img2threejs}
}

@misc{hu2024scenecraft,
  title         = {{SceneCraft}: An {LLM} Agent for Synthesizing {3D} Scene as {Blender} Code},
  author        = {Hu, Ziniu and Iscen, Ahmet and Jain, Aashi and Kipf, Thomas and Yue, Yisong and Ross, David A. and Schmid, Cordelia and Fathi, Alireza},
  year          = {2024},
  eprint        = {2403.01248},
  archivePrefix = {arXiv},
  primaryClass  = {cs.CV},
  url           = {https://arxiv.org/abs/2403.01248}
}

@inproceedings{raistrick2023infinigen,
  title     = {Infinite Photorealistic Worlds Using Procedural Generation},
  author    = {Raistrick, Alexander and Lipson, Lahav and Ma, Zeyu and Mei, Lingjie and Wang, Mingzhe and Zuo, Yiming and Kayan, Karhan and Wen, Hongyu and Han, Beining and Wang, Yihan and Newell, Alejandro and Law, Hei and Goyal, Ankit and Yang, Kaiyu and Deng, Jia},
  booktitle = {Proceedings of the IEEE/CVF Conference on Computer Vision and Pattern Recognition (CVPR)},
  month     = jun,
  year      = {2023},
  pages     = {12630--12641},
  url       = {https://openaccess.thecvf.com/content/CVPR2023/html/Raistrick_Infinite_Photorealistic_Worlds_Using_Procedural_Generation_CVPR_2023_paper.html}
}

@misc{guo2026worldclaw,
  title         = {{WorldClaw}: Agentic {3D} Open-World Generation at Scale},
  author        = {Guo, Chunchao and Li, Jinpeng and Li, Yang and Huang, Zilong},
  year          = {2026},
  eprint        = {2608.05248},
  archivePrefix = {arXiv},
  primaryClass  = {cs.AI},
  url           = {https://arxiv.org/abs/2608.05248}
}

@misc{feng2026funcroom,
  title         = {{FuncRoom-Agent}: Sequential Feed-Forward {3D} Functional Indoor Scene Generation},
  author        = {Feng, Hao and Zuo, Zhi and Liang, MingJian and Hu, Jingyu and Hu, Xiaowei and Wu, Liupengfei and Zhang, Dian and Fang, Guoxin and Liu, Zhengzhe},
  year          = {2026},
  eprint        = {2608.29519},
  archivePrefix = {arXiv},
  primaryClass  = {cs.CV},
  url           = {https://arxiv.org/abs/2608.29519}
}

@misc{zhang2025rlm,
  title         = {Recursive Language Models},
  author        = {Zhang, Alex L. and Kraska, Tim and Khattab, Omar},
  year          = {2025},
  eprint        = {2512.24601},
  archivePrefix = {arXiv},
  primaryClass  = {cs.AI},
  note          = {Version 3, revised May 11, 2026},
  url           = {https://arxiv.org/abs/2512.24601}
}

@inproceedings{zhang2018lpips,
  title     = {The Unreasonable Effectiveness of Deep Features as a Perceptual Metric},
  author    = {Zhang, Richard and Isola, Phillip and Efros, Alexei A. and Shechtman, Eli and Wang, Oliver},
  booktitle = {Proceedings of the IEEE Conference on Computer Vision and Pattern Recognition (CVPR)},
  month     = jun,
  year      = {2018},
  pages     = {586--595},
  url       = {https://openaccess.thecvf.com/content_cvpr_2018/html/Zhang_The_Unreasonable_Effectiveness_CVPR_2018_paper.html}
}

@misc{janachumi2017isometric,
  title        = {Isometric city},
  author       = {{JanaChumi}},
  year         = {2017},
  howpublished = {OpenGameArt.org asset pack, {CC0 1.0}},
  note         = {Published December 9, 2017; pack contains roads, vegetation, a fountain, a lake, civic buildings, other buildings, houses, and a shop},
  url          = {https://opengameart.org/content/isometric-city-0}
}
\bibliographystyle{plainnat}
\clearpage
\appendix
\section{Reproducibility Details}
\label{app:repro}

This appendix provides the solver instruction and the execution
settings used in the reported experiments.

\paragraph{Execution settings.}
Each solver call receives a reference crop, a matched camera
specification, a task brief, and the inherited parent snapshot.
When the agent requests children, the runner starts their solver
calls and resumes the parent after their components are returned.
The runner allows five levels in total. At this limit, the
agent completes the current component through local refinement.
The fixed two-level ablation sets the maximum two level.

\subsection{The solver instruction}
\label{app:solver}

The instruction below is used at every node in the main reconstruction
experiments. The node name,
depth, and directory paths are filled in for each call. Its inputs
are \code{target.png}, \code{view.json}, \code{brief.md}, and a
read-only \code{manifest.json} describing the inherited reference,
camera, and parent snapshot.

%The ablations in Section~\ref{sec:controls} use an earlier instruction revision without the explicit descent test in step~2. The two recursive variants in that study share this earlier instruction and differ in their depth limit.

% Keep the original tcolorbox containing the experimental solver
% instruction here, without rewriting its contents.

\begin{tcolorbox}[breakable,enhanced,colback=black!3,colframe=black!70,boxrule=0.6pt,arc=1.5pt,left=6pt,right=6pt,top=5pt,bottom=5pt,title={Solver instruction (released prompt, verbatim)},fonttitle=\bfseries\small]\small
\textbf{solve(node) --- the one recursive solver, at depth $d$.} This is an approved work order. Make reversible choices on your own recommendation; do not stop for approval.
Chain directory: \code{chain}; your node directory: \code{chain/fractal/node/}. Materials: \code{target.png} (this level's target image) / \code{view.json} (the viewport) / \code{brief.md}; \code{manifest.json} records the reference, camera and parent-snapshot versions you inherited (read-only). Camera contract: \code{chain/camera-contract.json} (read-only; if it does not exist and this is the root, solve and calibrate one first and verify the full-frame overlay by eye before locking it).

\emph{What you do at this level (one loop, identical at every level).}
(1)~\textbf{Whole}: render this level's viewport and put it side by side with \code{target.png} at matched magnification; look for the conspicuous residuals by eye; repeat as needed. Also take one or two rotated views --- any flatness a side view exposes (paper walls, billboard trees, parts without thickness) is a real residual and goes on the repair list right away.
(2)~\textbf{Parts}: repair here what can be repaired here. \textbf{A sub-problem that deserves its own solve} (one sub-object or sub-assembly that patching at this level cannot fix and that needs its own target image and loop) is not to be forced through. The test: take each part of the side-by-side where you can still see a difference and magnify it on its own. If what you see is \textbf{a group of objects and the layout relations between them} (arrangement, spacing, orientation, attachment) that this level has not yet produced, or \textbf{a single thing whose internal structure is complex enough to need its own loop}, it is a sub-problem: descend. If it is only a simple single thing or scattered small differences, finish it at this level. A level usually cuts two to four children; most scenes close in two or three levels. For a sub-problem, prepare its materials under \code{chain/fractal/<child>/} (\code{target.png} cut from your target and magnified / \code{view.json} / \code{brief.md}), list the child in \code{children.json} in your directory (a JSON array, e.g.\ \code{["part-a","part-b"]}), then \textbf{end the session} --- the runner starts this same solver for every child and wakes you with the results when they are done.
(3)~\textbf{Whole again} (after being woken): the children's \code{part.json} files are in place; integrate them into your level, go back to the side-by-side of the whole at this level and settle continuity and relations; if more descent is needed, update \code{children.json} and end again, otherwise finish.
(4)~\textbf{Finish}: write \code{chain/fractal/node/part.json} (this level's final component, with child references) + \code{account.md} (this level's round-by-round account). \code{part.json} on disk means this level is done.

\emph{Completion (at every level).} First identify: what this is, in what style, and what a complete instance of its kind looks like --- imagine the whole thing in your mind, then calibrate that mental image against the visible evidence in the reference (two or three sentences in \code{account.md}). Then: what is visible, match to the reference pixel by pixel; what is not visible (back faces, occluded parts, boundary continuations), complete from the calibrated mental image with world knowledge in the same style --- a house has four walls and a full roof, a tree crown goes all the way round, a road leads somewhere, terrain continues to the boundary. Self-check: render the scene from the four compass directions and look; stage-set feel and flatness (empty backs, dead-end roads, floating objects, paper parts) must be gone before completion counts; re-verify the visible region against the reference in the same window and do not let the completion pull it away.

Discipline: eyes decide, metrics are only recorded; free-form boxes; matched magnification; no git; a child's write scope is its own \code{fractal/<child>/}. Everything else about the craft is yours to decide. Begin.
\end{tcolorbox}

The model retains discretion over geometry, materials, decomposition, and repair. Resource limits, schema checks, and evidence recording are automatic runtime responsibilities. The instruction neither authorizes external actions nor uses a generated appearance score as a success gate.

\section{Design Philosophy}
\label{app:argument}

Complex-world reconstruction requires both detailed work on individual
components and repeated checks of how those components fit together.
The design in Section~\ref{sec:method} organizes these tasks around a
complete reconstruction process that can be applied at every scale.

\paragraph{A complete solve for each subworld.}
A subworld has its own reference view, editable code, and reconstruction
loop. A facade, for example, needs an arrangement of windows and signs
as well as a coherent placement within its building. Giving the facade
a complete solve allows the agent to establish that arrangement,
reconstruct its details through further calls, and inspect the assembled
result. The same procedure applies to a building, an interacting assembly,
or a terrain region. The agent chooses the decomposition from the visible
structure and the remaining work. Examples in the instruction illustrate
modeling choices; the recursive procedure remains applicable to other
components and scene types.

\paragraph{A shared projection for local inspection.}
A child needs a target that preserves its relationship to the surrounding
scene. We therefore pair each reference crop with the corresponding
camera viewport, inheriting the parent's camera pose and applying the
same crop coordinates and magnification. This gives local comparisons
a consistent frame for judging shape, scale, placement, and occlusion.
The crop focuses inspection on evidence already present in the input.
When the root camera or a shared scene dependency changes, affected
child views are refreshed before further work.

\paragraph{Trusting visual judgment.}
We rely on the model's visual understanding and reasoning to guide
reconstruction. The agent examines the reference and current render
at matched magnification, identifies discrepancies in shape, placement,
and spatial relationships, and decides how to address them. A missing
window may call for a local edit; a complex facade may need its own
complete solve. After edits and child returns, the agent looks again
to assess the result and determine the next step. The recursive
structure supports this judgment by giving the model focused views
of unresolved details and repeated opportunities to inspect how the
parts fit together. Numerical appearance metrics are recorded for
evaluation; the model's visual comparisons guide reconstruction
and stopping.

\paragraph{Completing unobserved structure.}
A single image leaves hidden surfaces and out-of-frame continuations
ambiguous. Our completion instruction asks the agent to use its knowledge
of the depicted objects and style when constructing such geometry.
These additions express assumptions about the unobserved scene. Rotated
views help inspect thickness, support, and continuity, and the resulting
program is checked again from the reference view to identify changes in
visible silhouettes, occlusions, or shading. We assess visible fidelity
against the reference and use the other views to examine the geometry
produced by these assumptions.

\paragraph{Refining relationships after composition.}
Local reviews cover the structures and relationships they actually
inspect. A review of building shape and facade detail can leave the
entrance's orientation toward a shared street unexamined. Changes made
by separate child calls can also affect each other: two components edited
against earlier surroundings may overlap after composition. The parent
therefore renders the assembled result and revisits placement, contact,
occlusion, and shared geometry. It can edit these relationships directly
or reopen affected child tasks. Shared errors are repaired at the ancestor
that controls their cause. Compatible children may run concurrently,
followed by this same joint refinement step. The purpose of returning is
to evaluate and improve the composition produced by the child calls.

\paragraph{Code as the world representation.}
We represent the world as executable code that describes how its
geometry, materials, and spatial arrangement are constructed.
The code exposes both the procedures that generate each component
and the parameters that control its properties. This gives the
agent concrete ways to act on visual observations: change a
dimension, revise a modeling procedure, or adjust how components
are placed together. It also supports recursive construction.
Each child returns a subprogram that the parent can compose into
the world and continue to edit. For example, a parent can reposition
a reconstructed building while retaining the code that generates
its windows and roof. The completed world preserves these
procedures and component references, supporting further editing,
reuse, and rendering from new viewpoints.
\end{document}